# Optimal Schedules for Parallelizing Anytime Algorithms: The Case of Shared Resources

**Lev Finkelstein**                              LEV@CS.TECHNION.AC.IL
**Shaul Markovitch**                             SHAULM@CS.TECHNION.AC.IL
**Ehud Rivlin**                                  EHUDR@CS.TECHNION.AC.IL
*Computer Science Department*
*Technion - Israel Institute of Technology*
*Haifa 32000, Israel*

## Abstract

The performance of anytime algorithms can be improved by simultaneously solving several instances of algorithm-problem pairs. These pairs may include different instances of a problem (such as starting from a different initial state), different algorithms (if several alternatives exist), or several runs of the same algorithm (for non-deterministic algorithms). In this paper we present a methodology for designing an optimal scheduling policy based on the statistical characteristics of the algorithms involved. We formally analyze the case where the processes share resources (a single-processor model), and provide an algorithm for optimal scheduling. We analyze, theoretically and empirically, the behavior of our scheduling algorithm for various distribution types. Finally, we present empirical results of applying our scheduling algorithm to the Latin Square problem.

## 1. Introduction

Assume that our task is to learn a concept with a predefined success rate, measured on a given test set. Assume that we can use two alternative learning algorithms, one which learns fast but requires some preprocessing, and another which works more slowly but requires no preprocessing. Can we possibly benefit from using *both* learning algorithms in parallel to solve *one* learning task on a single-processor machine?

Another area of application is that of constraint satisfaction problems. Assume that a student tries to decide between two elective courses by trying to schedule each of them with the set of her compulsory courses. Should the student try to solve the two sets of constraints sequentially or should the two computations be somehow interleaved?

Assume now that a crawler searches for a specific page in a site. If we had more than one starting point, the process could be speeded up by simultaneous application of the crawler from a few (or all) of them. However, what would be the optimal strategy if the bandwidth were restricted?

What do the above examples have in common?

- There are potential benefits to be gained from the uncertainty in the amount of resources that will be required to solve more than one instance of the algorithm-problem pair. We can use different algorithms (in the first example) and different problems (in the last two examples). For non-deterministic algorithms, we can also use different runs of the same algorithm.

- Each process is executed with the purpose of satisfying a given goal predicate. The task is considered accomplished when one of the runs succeeds.

- If the goal predicate is satisfied at time $t^*$, then it is also satisfied at any time $t > t^*$. This property is equivalent to utility monotonicity of *anytime algorithms* (Dean & Boddy, 1988; Horvitz, 1987), where solution quality is restricted to Boolean values.

Our objective is to provide a schedule that minimizes the expected cost, possibly under some constraints (for example, processes may share resources). Such problem definition is typical for *rational-bounded* reasoning (Simon, 1982; Russell & Wefald, 1991). This problem resembles those faced by *contract algorithms* (Russell & Zilberstein, 1991; Zilberstein, 1993). There, given the allocated resources, the task is to construct an algorithm providing a solution of the highest quality. In our case, given quality requirements, the task is to construct an algorithm that solves the problem using minimal resources.

There are several research works that deal with similar problems. Simple parallelization, with no information exchange between the processes, may speed up the process due to high diversity in solution times. For example, Knight (1993) showed that using many reactive agents employing RTA* search (Korf, 1990) is more beneficial than using a single deliberative agent. Another example is the work of Yokoo and Kitamura (1996), who used several search agents in parallel, with agent rearrangement after preallotted periods of time. Janakiram, Agrawal, and Mehrotra (1988) showed that for many common distributions of solution time, simple parallelization leads to at most linear speedup. One exception is the family of *heavy-tailed* distributions (Gomes, Selman, & Kautz, 1998) for which it is possible to obtain superlinear speedup by simple parallelization.

A superlinear speedup can also be obtained when we have access to the internal structure of the processes involved. For example, Clearwater, Hogg, and Huberman (1992) reported superlinear speedup for cryptarithmetic problems as a result of information exchange between the processes. Another example is the works of Kumar and Rao (Rao & Kumar, 1987; Kumar & Rao, 1987; Rao & Kumar, 1993), devoted to parallelizing standard search algorithms, where superlinear speedup is obtained by dividing the search space.

An interesting domain-independent approach is based on "portfolio" construction (Huberman, Lukose, & Hogg, 1997; Gomes & Selman, 1997). In this approach, a different amount of resources is allotted to each process. This can reduce both expected resource consumption and its variance.

In the case of non-deterministic algorithms, another way to benefit from solution time diversity is to restart the same algorithm in attempt to switch to a better trajectory. Such a framework was analyzed in detail by Luby, Sinclair, and Zuckerman (1993) for the case of a single processor and by Luby and Ertel (1994) for the multiprocessor case. In particular, it was proven that for a single processor, the optimal strategy is to periodically restart the algorithm after a constant amount of time until the solution is found. This strategy was successfully applied to combinatorial search problems by Gomes, Selman, and Kautz (1998).

There are several settings, however, where the restart strategy is not optimal. If the goal is to schedule a number of runs of a single non-deterministic algorithm, such that this number is limited due to the nature of the problem (for example, robotic search), the restart strategy is applicable but not optimal. A special case of the above settings is scheduling a number of runs of a deterministic algorithm with a finite set of available initial configurations (inputs). Finally, the case where the goal is to schedule a set of algorithms different from each other is out of the scope of the restart strategy.

The goal of this research is to develop a methodology for designing an optimal scheduling policy for any number of instances of algorithm-problem pairs, where the algorithms can be either deterministic or non-deterministic. We present a formal framework for scheduling parallel anytime algorithms for the case where the processes share resources (a single-processor model), based on the statistical characteristics of the algorithms involved. The framework assumes that we know the probability of the goal condition to be satisfied as a function of time (a *performance profile* (Simon, 1955; Boddy & Dean, 1994) restricted to Boolean quality values). We analyze the properties of optimal schedules for the suspend-resume model, where allocation of resources is performed on mutual exclusion basis, and show that in most cases an extension of the framework to intensity control, where resources may be allocated simultaneously and proportionately to multiple demands, does not yield better schedules. We also present an algorithm for building optimal schedules. Finally, we demonstrate experimental results for the optimal schedules.

## 2. Motivation

Before starting the formal discussion, we would like to illustrate how different scheduling strategies can affect the performance of a system of two search processes. The first example has a very simple setup which allows us to perform a full analysis. In the second example, we show quantitative results for a real CSP problem.

### 2.1 Scheduling DFS Search Processes

Assume DFS with random tie-breaking is applied to a simple search space shown in Figure 1, but that only two runs of the algorithm are allowed[1]. There is a very large number of paths to the goal, half of them of length 10, quarter of them of length 40, and quarter of them of length 160. When one of the processes finds the solution, the task is considered accomplished.

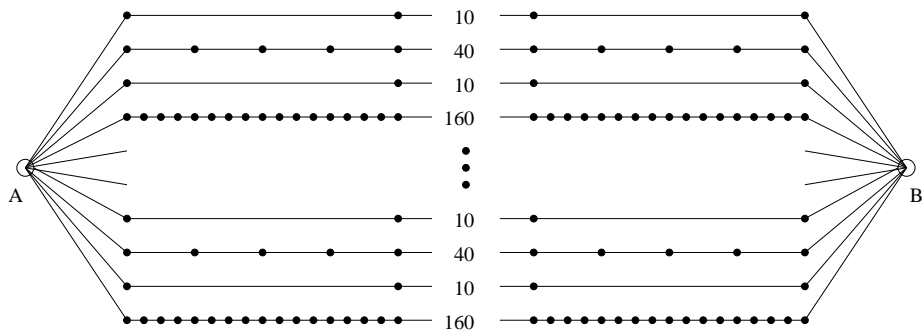

Figure 1: A simple search task: two DFS-based agents search for a path from $A$ to $B$. Scheduling the processes may reduce costs.

---

1. Such a limit can follow, for example, from physical constraints, such as for the problem of robotic search. For unlimited number of runs the optimal results would be provided by the restart strategy.

We consider a single-processor system, where the two processes cannot run simultaneously. Let us denote the processes by $A_1$ and $A_2$, and by $L_1$ and $L_2$ the actual path lengths for $A_1$ and $A_2$ respectively for the particular run.

The application of a single processes (without loss of generality, $A_1$) gives us the expected execution time of $1/2 \times 10 + 1/4 \times 40 + 1/4 \times 160 = 55$, as is shown in Figure 2.

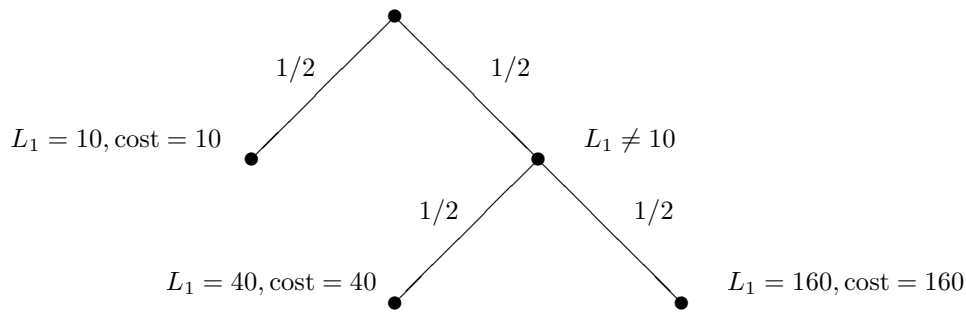

Figure 2: Path lengths, probabilities and costs for running a single process

We can improve the performance by simulating a simultaneous execution of two processes. For this purpose, we allow each of the processes to expand a *single* node, and to switch to the other process (without loss of generality, $A_1$ starts first). In this case, the expected execution time is $1/2 \times 19 + 1/4 \times 20 + 1/8 \times 79 + 1/16 \times 80 + 1/16 \times 319 = 49.3125$, as is shown in Figure 3.

Finally, if we know the distribution of path lengths, we can allow $A_1$ to open 10 nodes; if $A_1$ fails, we can stop it and allow $A_2$ to open 10 nodes; if $A_2$ fails as well, we can allow $A_1$ to open the next 30 nodes, and so forth. In this scenario, $A_1$ and $A_2$ switch after 10 and 40 nodes (if both processes fail to find a solution after 40 nodes, it is guaranteed to be found by $A_1$ after 160 nodes). This scheme is shown in Figure 4, and the expected time is $1/2 \times 10 + 1/4 \times 20 + 1/8 \times 50 + 1/16 \times 80 + 1/16 \times 200 = 33.75$.

## 2.2 The Latin Square Example

The task in the Latin Square problem is to place $N$ symbols on an $N \times N$ square such that each symbol appears only once in each row and each column. An example is shown in Figure 5.

A more interesting problem arises when the square is partially filled. The problem in this case may be solvable (see the left side of Figure 6) or unsolvable (see the right side of Figure 6). The problem of satisfiability of a partially filled Latin Square is a typical constraint-satisfaction problem. We consider a slight variation of this task. Let us assume that *two* partially filled squares are available, and we need to decide whether at least one of them is solvable. We assume that we are allocated a single processor. We attempt to speed up the time of finding a solution by starting to solve the two problems from two different initial configurations in parallel.

Each of the processes employs a deterministic heuristic DFS with the First-Fail heuristic (Gomes & Selman, 1997). We consider 10%-filled $20 \times 20$ Latin Squares. The behavior of a single process measured on a set of 50,000 randomly generated samples is shown in each run is limited to 25,000 search steps (only 88.6% of the problems are solvable under this condition). If we apply the algorithm only on one of the available two initial configurations, the average number of search steps is 3777. If we run two processes in parallel (alternating after each step), we obtain a result of 1358 steps. If we allow a single switch at the *optimal* point (an analogue of the restart technique (Luby et al., 1993; Gomes et al., 1998) for two processes), we get 1376 steps on average (the optimal point is after 1311 steps). Finally, if we interleave the processes, switching at the points corresponding to 679, 3072, and 10208 of total steps, the average number of steps is 1177. The above results were averaged over a test set of 25,000 pairs of initial configurations.

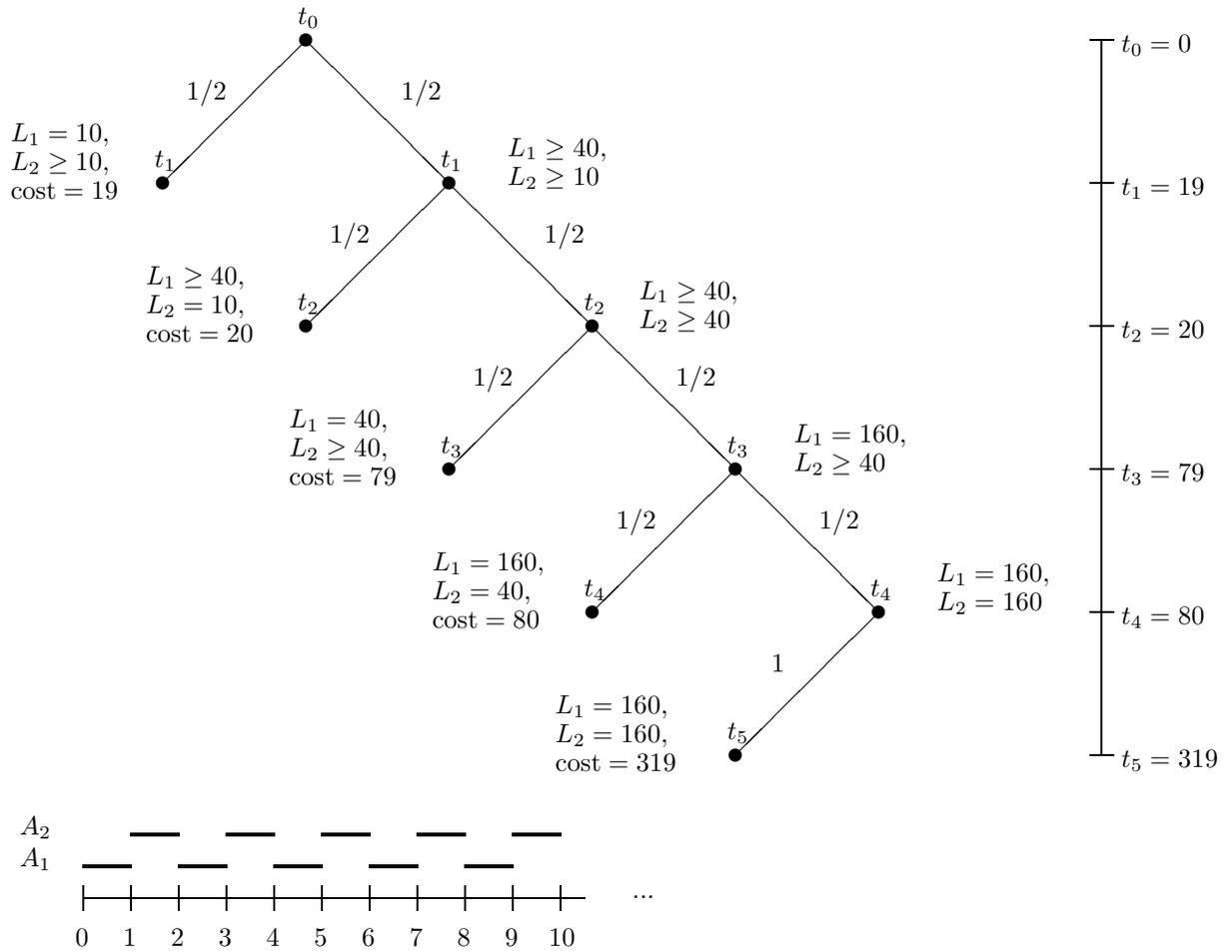

Figure 3: Path lengths, probabilities and costs for simulating a simultaneous execution

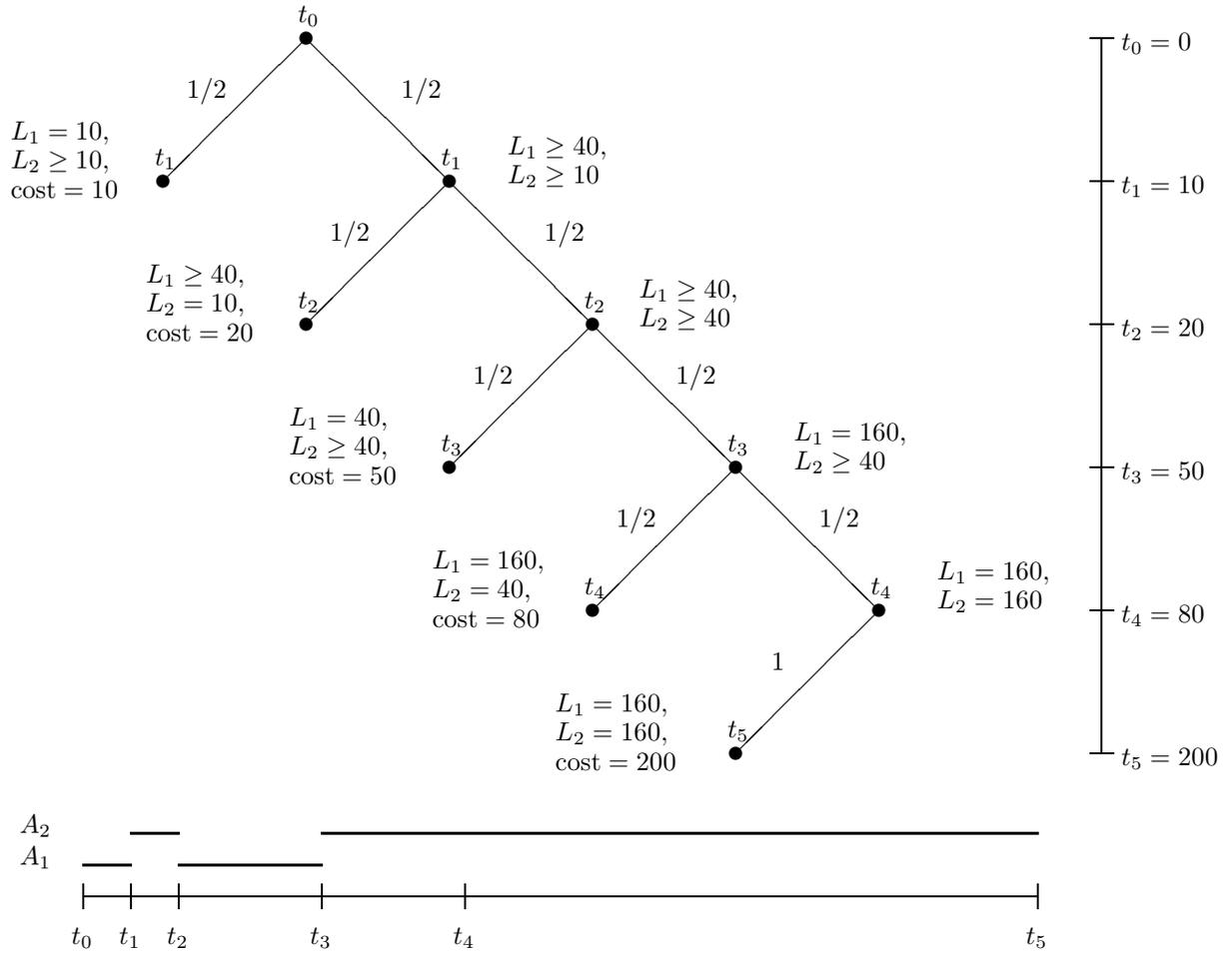

Figure 4: Path lengths, probabilities and costs for the interleaved execution

| 1 | 2 | 3 | 4 | 5 |
|---|---|---|---|---|
| 3 | 4 | 5 | 1 | 2 |
| 5 | 1 | 2 | 3 | 4 |
| 2 | 3 | 4 | 5 | 1 |
| 4 | 5 | 1 | 2 | 3 |

Figure 5: An example of a $5 \times 5$ Latin Square.

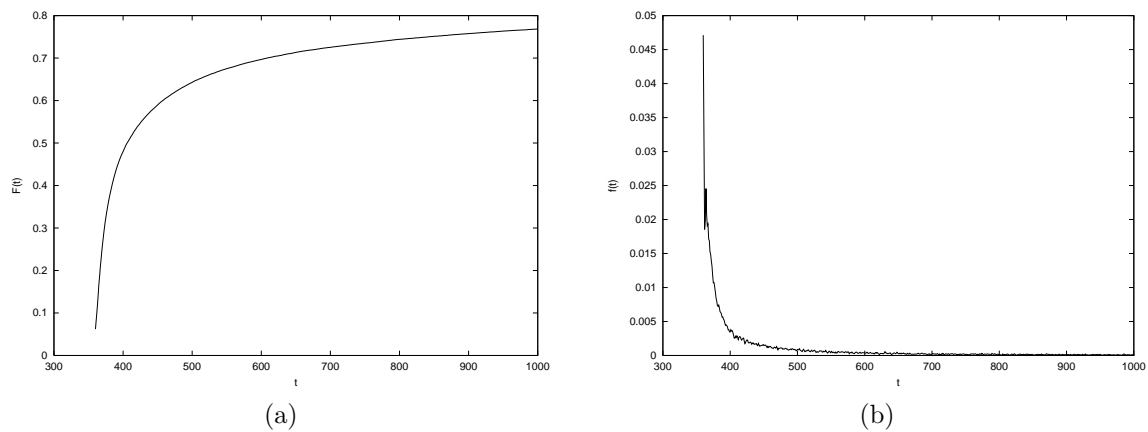

Figure 6: An example of solvable (to the left) and unsolvable (to the right) prefilled 5 × 5 Latin Squares.

Figure 7. Figure 7(a) shows the probability of finding a solution as a function of the number of search steps, and Figure 7(b) shows the corresponding distribution density. Assume that

Figure 7: The behavior of DFS with the First-Fail heuristic on 10%-filled 20 × 20 Latin Squares. (a) The probability of finding a solution as a function of the number of search steps; (b) The corresponding distribution density.

The last sequence of switch points is an optimal schedule for the process with behavior described by the graphs in Figure 7. In the rest of the paper we present an algorithm for deriving such optimal schedules.

## 3. A Framework for Parallelization Scheduling

In this section we formalize the intuitive description of parallelization scheduling. The first part of this framework is similar to our framework presented in (Finkelstein & Markovitch, 2001).

Let $\mathcal{S}$ be a set of states, $t$ be a time variable with non-negative real values, and $\mathcal{A}$ be a random process such that each realization (trajectory) $A(t)$ of $\mathcal{A}$ represents a mapping from $\mathcal{R}^+$ to $\mathcal{S}$. Let $X_0$ be a random variable defined over $\mathcal{S}$. Since an algorithm $Alg$ starting from an initial state $S_0$ corresponds to a single trajectory (for deterministic algorithms), or to a set of trajectories with an associated distribution (for non-deterministic algorithms), the pair $\langle X_0, Alg \rangle$, where $X_0$ stands for the initial state, can be viewed as a random process. Drawing a trajectory for such a process corresponds, without loss of generality, to a two-step procedure: first an initial state $S_0$ is drawn for $X_0$, and then a trajectory $A(t)$ starting from $S_0$ is drawn for $Alg$. Thus, the source of randomness is either the randomness of the initial state, or the randomness of the algorithm (which can come from the algorithm itself or from the environment), or both.

Let $\mathcal{S}^* \subseteq \mathcal{S}$ be a designated set of states, and $G : \mathcal{S} \to \{0, 1\}$ be the characteristic function of $\mathcal{S}^*$ called the *goal predicate*. The behavior of a trajectory $A(t)$ of $\mathcal{A}$ with respect to the goal predicate $G$ can be written as $G(A(t))$, which we denote by $\widehat{G_A}(t)$. We say that $\mathcal{A}$ is *monotonic* over $G$ if and only if $\widehat{G_A}(t)$ is a non-decreasing function for each trajectory $A(t)$ of $\mathcal{A}$. Under the above assumptions $\widehat{G_A}(t)$ is a step function with at most one discontinuity point.

Let $\mathcal{A}$ be monotonic over $G$. From the definitions above we can see that the behavior of $G$ for each trajectory $A(t)$ of $\mathcal{A}$ can be described by a single point $\widehat{t}_{A,G}$, the first point after which the goal predicate is true, i.e, $\widehat{t}_{A,G} = \inf_t \{t | \widehat{G_A}(t) = 1\}$. If $\widehat{G_A}(t)$ is always 0, we say that $\widehat{t}_{A,G}$ is not defined. Therefore, we can define a random variable, which for each trajectory $A(t)$ of $\mathcal{A}$ with $\widehat{t}_{A,G}$ defined, corresponds to $\widehat{t}_{A,G}$. The behavior of this variable can be described by its distribution function $F(t)$. At the points where $F(t)$ is differentiable, we use the probability density $f(t) = F'(t)$.

It is important to note that in practice not every trajectory of $\mathcal{A}$ leads to the goal predicate satisfaction even after infinitely large time. That means that the set of the trajectories where $\widehat{t}_{A,G}$ is undefined is not necessarily of measure zero. That is why we define the *probability of success* $p$ as the probability of $A(t)$ to have $\widehat{t}_{A,G}$ defined[2]. For the Latin Square example described in Section 2.2, the probability of success is 0.886, and the graphs in Figure 7 correspond to $pF(t)$ and $pf(t)$.

Assume now that we have a system of $n$ random processes $\mathcal{A}_1, \ldots \mathcal{A}_n$ with corresponding distribution functions $F_1, \ldots, F_n$ and goal predicates $G_1, \ldots, G_n$. If the distribution functions $F_i$ and $F_j$ are identical, we refer to $\mathcal{A}_i$ and $\mathcal{A}_j$ as *F-equivalent*.

We define a *schedule* of the system as a set of binary functions $\{\theta_i\}$, where at each moment $t$, the $i$-th process is active if $\theta_i(t) = 1$ and idle otherwise. We refer to this scheme as *suspend-resume* scheduling. A possible generalization of this framework is to extend the suspend/resume control to a more refined mechanism that allows us to determine the

---

2. Another way to express the possibility that a process will not reach a goal state is to use $F(t)$ that approach $1 - p$ when $t \to \infty$. We prefer to use $p$ explicitly because the distribution function must meet the requirement $\lim_{t \to \infty} F(t) = 1$.

*intensity* with which each process acts. For software processes, this means varying the fraction of CPU utilization; for tasks like robot navigation this implies changing the speed of the robots. Mathematically, using intensity control is equivalent to replacing the binary functions $\theta_i(t)$ with continuous functions with a range between zero and one[3].

Note that scheduling makes the term *time* ambiguous. On one hand, we have the *subjective* time for each process, consumed only when the process is active. This kind of time corresponds to some resource consumed by the process. On the other hand, we have an *objective* time measured from the point of view of an external observer. The distribution function $F_i(t)$ of each process is defined over its subjective time, while the cost function (see below) may use both kinds of times. Since we are using several processes, all the formulas in this paper are based on the objective time.

Let us denote by $\sigma_i(t)$ the total time that process $i$ has been active before $t$. By definition,

$$\sigma_i(t) = \int_0^t \theta_i(x)dx. \tag{1}$$

In practice $\sigma_i(t)$ provides the mapping from the objective time $t$ to the subjective time of the $i$-th process, and we refer to these functions as *subjective schedule functions*. Since $\theta_i$ can be obtained from $\sigma_i$ by differentiation, we often describe schedules by $\{\sigma_i\}$ instead of $\{\theta_i\}$.

The processes $\{\mathcal{A}_i\}$ with goal predicates $\{G_i\}$ running under schedules $\{\sigma_i\}$ result in a new process $\mathcal{A}$, with a goal predicate $G$. $G$ is the disjunction of $G_i$ ($G(t) = \bigvee_i G_i(t)$), and therefore $\mathcal{A}$ is monotonic over $G$. We denote the distribution function of the corresponding random variable by $F_n(t, \sigma_1, \ldots, \sigma_n)$, and the corresponding distribution density by $f_n(t, \sigma_1, \ldots, \sigma_n)$.

Assume that we are given a monotonic non-decreasing *cost* function $u(t, t_1, \ldots, t_n)$, which depends on the objective time $t$ and the subjective times per process $t_i$. We also assume that $u(0, t_1, \ldots, t_n) = 0$. Since the subjective times can be calculated by $\sigma_i(t)$, we actually have $u = u(t, \sigma_1(t), \ldots, \sigma_n(t))$.

The *expected* cost of schedule $\{\sigma_i\}$ can be expressed, therefore, as[4]

$$E_u(\sigma_1, \ldots, \sigma_n) = \int_0^{+\infty} u(t, \sigma_1, \ldots, \sigma_n)f_n(t, \sigma_1, \ldots, \sigma_n)dt \tag{2}$$

(for the sake of readability, we omit $t$ in $\sigma_i(t)$). Under the suspend-resume model assumptions, $\sigma_i$ must be differentiable (except for a countable set of process switch points) and have derivatives of 0 or 1 that would ensure correct values for $\theta_i$. Under intensity control assumptions, the derivatives of $\sigma_i$ must lie between 0 and 1.

We consider two alternative setups for resource sharing between the processes:

1. The processes share resources on a mutual exclusion basis. That means that exactly one process can be active at each moment, and the processes will be active one after another until the goal is reached by one of them. In this case the sum of derivatives of $\sigma_i$ is always one[5]. The case of shared resources corresponds to the case of several processes running on a single processor.

---

3. A special case of such a setup using constant intensities was described by Huberman, Lukose, and Hogg (1997).

4. The generalization to the case where the probability of success $p$ is not 1 is considered at the end of the next section.

2. The processes are fully independent: there are no additional constraints on $\sigma_i$. This case corresponds to $n$ independent processes running on $n$ processors.

Our goal is to find a schedule which minimizes the expected cost (2) under the corresponding constraints. The current paper is devoted to the case of shared processes. The case of independent resources was studied in (Finkelstein, Markovitch, & Rivlin, 2002).

The scheduled algorithms considered in this framework can be viewed as anytime algorithms. The behavior of anytime algorithms is usually characterized by their performance profile – the expected quality of the algorithm output as a function of the alloted resources. The goal predicate $G$ can be viewed as a quality function with two possible values, and thus the distribution function $F(t)$ meets the definition of performance profile, where time plays the role of resource.

## 4. Suspend-Resume Based Scheduling

In this section we consider the case of suspend-resume based control ($\sigma_i$ are continuous functions with derivatives 0 or 1).

**Claim 1** *The expressions for the goal-time distribution $F_n(t, \sigma_1, \ldots, \sigma_n)$ and the expected cost $E_u(\sigma_1, \ldots, \sigma_n)$ are as follows[6]:*

$$F_n(t, \sigma_1, \ldots, \sigma_n) = 1 - \prod_{i=1}^{n}(1 - F_i(\sigma_i)), \tag{3}$$

$$E_u(\sigma_1, \ldots, \sigma_n) = \int_0^{+\infty} \left( u_t' + \sum_{i=1}^{n} \sigma_i' u_{\sigma_i}' \right) \prod_{i=1}^{n}(1 - F_i(\sigma_i)) dt. \tag{4}$$

**Proof:** Let $t_i$ be the time it would take the $i$-th process to meet the goal if acted alone (if the process fails to reach the goal, we consider $t_i = \infty$). Let $t^*$ be the time it takes the system of $n$ processes to reach the goal. In this case, $t^*$ is distributed according to $F_n(t, \sigma_1, \ldots, \sigma_n)$, and $t_i$ are distributed according to $F_i(t)$. Thus, because the processes, given a schedule, are independent, we obtain

$$F_n(t, \sigma_1, \ldots, \sigma_n) = P(t^* \leq t) = 1 - P(t^* > t) = 1 - P(t_1 > \sigma_1(t)) \times \ldots \times P(t_n > \sigma_n(t)) =$$

$$1 - (1 - F_1(\sigma_1(t))) \times \ldots \times (1 - F_n(\sigma_n(t))) = 1 - \prod_{i=1}^{n}(1 - F_i(\sigma_i(t))),$$

which corresponds to (3). Since $F(t)$ is a distribution over time, we assume $F(t) = 0$ for $t \leq 0$.

---

5. This fact is obvious for the case of suspend-resume control, and for intensity control it is reflected in Lemma 3.
6. $u_t'$ and $u_{\sigma_i}'$ stand for partial derivatives of $u$ by $t$ and by $\sigma_i$ respectively.

The average cost function will therefore be

$$E_u(\sigma_1, \ldots, \sigma_n) = \int_0^{+\infty} u(t, \sigma_1, \ldots, \sigma_n) f_n(t, \sigma_1, \ldots, \sigma_n) dt =$$

$$- \int_0^{+\infty} u(t, \sigma_1, \ldots, \sigma_n) d(1 - F_n(t, \sigma_1, \ldots, \sigma_n)) =$$

$$- u(t, \sigma_1, \ldots, \sigma_n)(1 - F_n(t, \sigma_1, \ldots, \sigma_n))|_0^\infty + \int_0^{+\infty} \frac{du(t, \sigma_1, \ldots, \sigma_n)}{dt} \prod_{i=1}^n (1 - F_i(\sigma_i)) dt.$$

Since $u(0, \sigma_1, \ldots, \sigma_n) = 0$ and $F_n(\infty, \sigma_1, \ldots, \sigma_n) = 1$, the first term in the last expression is 0. Besides, since the full derivative of $u$ by $t$ can be written as

$$\frac{du(t, \sigma_1, \ldots, \sigma_n)}{dt} = u'_t + \sum_{i=1}^n \sigma'_i u'_{\sigma_i},$$

we obtain

$$E_u(\sigma_1, \ldots, \sigma_n) = \int_0^{+\infty} \left( u'_t + \sum_{i=1}^n \sigma'_i u'_{\sigma_i} \right) \prod_{i=1}^n (1 - F_i(\sigma_i)) dt,$$

which completes the proof.
*Q.E.D.*

Note that in the case of $\sigma_i(t) = t$ and $F_i(t) = F(t)$ for all $i$ (parallel application of $n$ $F$-equivalent processes), we obtain the formula presented in (Janakiram et al., 1988), i.e., $F_n(t) = 1 - (1 - F(t))^n$.

In the rest of this section we show a formal solution (necessary conditions and an algorithm) for the framework with shared resources. We start with two processes and present the formulas and the algorithm, and then generalize the solution for an arbitrary number of processes. For the case of two processes, we only assume that $u$ is differentiable.

For the more elaborated setup of $n$ processes, we assume that the total cost is a linear combination of the objective time and all the subjective times, and the subjective times are of the same weight:

$$u(t, \sigma_1, \ldots, \sigma_n) = at + b \sum_{i=1}^n \sigma_i(t). \tag{5}$$

Since time is consumed if and only if there is an active process, and the trivial case where all the processes are idle may be ignored, we obtain (without loss of generality)

$$E_u(\sigma_1, \ldots, \sigma_n) = \int_0^\infty \prod_{j=1}^n (1 - F_j(\sigma_j)) dt \to \min. \tag{6}$$

This assumption is made to keep the expressions more readable. The solution process remains the same for the general form of $u$.

## 4.1 Necessary Conditions for an Optimal Solution for Two Processes

Let $A_1$ and $A_2$ be two processes sharing a resource. While working, one process locks the resource, and the other is necessarily idle. We can show that such dependency yields a strong constraint on the behavior of the process, allowing the building of an effective algorithm for solving the minimization problem.

For the suspend-resume model, therefore, only two states of the system are possible: $A_1$ is active and $A_2$ is idle ($S_1$); and $A_1$ is idle and $A_2$ is active ($S_2$). We ignore the case where both processes are idle, since removing such a state from the schedule will not increase the cost. Therefore, the system continuously alternates between the two states: $S_1 \rightarrow S_2 \rightarrow S_1 \rightarrow S_2 \rightarrow \ldots$. We call the time interval corresponding to each pair $\langle S_1, S_2 \rangle$ a *phase* and denote phase $k$ by $\Phi_k$. If we denote the process switch points by $t_i$, the phase $\Phi_k$ corresponds to $[t_{2k-2}, t_{2k}]$. See Figure 8 for an illustration.

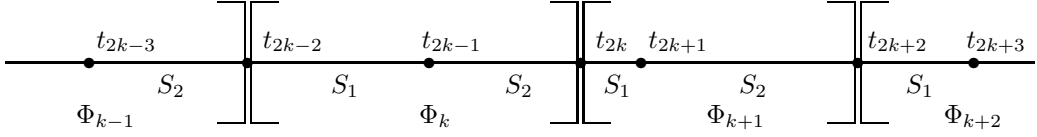

Figure 8: Notations for times, states and phases for two processes

By this scheme, $A_1$ is active in the intervals $[t_0, t_1]$, $[t_2, t_3]$, $\ldots$, $[t_{2k}, t_{2k+1}]$, $\ldots$, and $A_2$ is active in the intervals $[t_1, t_2]$, $[t_3, t_4]$, $\ldots$, $[t_{2k+1}, t_{2k+2}]$, $\ldots$.

Let us denote by $\zeta_{2k-1}$ the total time that $A_1$ has been active before $t_{2k-1}$, and by $\zeta_{2k}$ the total time that $A_2$ has been active before $t_{2k}$. By phase definition, $\zeta_{2k-1}$ and $\zeta_{2k}$ correspond to the cumulative time spent in phases 1 to $k$ in states $S_1$ and $S_2$ respectively. There exists a one-to-one correspondence between the sequences $\zeta_i$ and $t_i$:

$$\zeta_i + \zeta_{i+1} = t_{i+1}. \tag{7}$$

Moreover, by definition of $\zeta_i$ we have

$$\begin{aligned}
\sigma_1(t_{2k-1}) = \sigma_1(t_{2k}) = \zeta_{2k-1}, \\
\sigma_2(t_{2k}) = \sigma_2(t_{2k+1}) = \zeta_{2k}.
\end{aligned} \tag{8}$$

Under the process switch scheme as defined above, the subjective schedule functions $\sigma_1$ and $\sigma_2$ in time intervals $[t_{2k}, t_{2k+1}]$ (state $S_1$ of phase $\Phi_{k+1}$) have the form

$$\begin{aligned}
\sigma_1(t) = t - t_{2k} + \sigma_1(t_{2k}) = t - t_{2k} + \zeta_{2k-1} = t - \zeta_{2k}, \\
\sigma_2(t) = \sigma_2(t_{2k}) = \zeta_{2k}.
\end{aligned} \tag{9}$$

Similarly, in the intervals $[t_{2k+1}, t_{2k+2}]$ (state $S_2$ of phase $\Phi_{k+1}$), the subjective schedule functions are defined as

$$\begin{aligned}
\sigma_1(t) = \sigma_1(t_{2k+1}) = \zeta_{2k+1}, \\
\sigma_2(t) = t - t_{2k+1} + \sigma_2(t_{2k+1}) = t - t_{2k+1} + \zeta_{2k} = t - \zeta_{2k+1}.
\end{aligned} \tag{10}$$

Let us denote

$$v(t_1, t_2) = u'_t(t_1 + t_2, t_1, t_2) + u'_{\sigma_1}(t_1 + t_2, t_1, t_2) + u'_{\sigma_2}(t_1 + t_2, t_1, t_2)$$

and

$$v_i(t_1, t_2) = u'_t(t_1 + t_2, t_1, t_2) + u'_{\sigma_i}(t_1 + t_2, t_1, t_2).$$

To provide an optimal solution for the suspend/resume model, we may split (4) to phases $\Phi_k$ and write it as

$$E_u(\sigma_1, \ldots, \sigma_n) = \sum_{k=1}^{\infty} \int_{t_{2k-2}}^{t_{2k}} v(\sigma_1, \sigma_2)(1 - F_1(\sigma_1))(1 - F_2(\sigma_2))dt. \tag{11}$$

The last expression may be rewritten as

$$\begin{aligned}
E_u(\sigma_1, \ldots, \sigma_n) = \\
\sum_{k=0}^{\infty} \int_{t_{2k}}^{t_{2k+1}} v(\sigma_1, \sigma_2)(1 - F_1(\sigma_1))(1 - F_2(\sigma_2))dt + \\
\sum_{k=0}^{\infty} \int_{t_{2k+1}}^{t_{2k+2}} v(\sigma_1, \sigma_2)(1 - F_1(\sigma_1))(1 - F_2(\sigma_2))dt.
\end{aligned} \tag{12}$$

Using (9) on interval $[t_{2k}, t_{2k+1}]$, performing substitution $x = t - \zeta_{2k}$, and using (7), we obtain

$$\begin{aligned}
\int_{t_{2k}}^{t_{2k+1}} v(\sigma_1, \sigma_2)(1 - F_1(\sigma_1))(1 - F_2(\sigma_2))dt = \\
\int_{t_{2k}}^{t_{2k+1}} v_1(t - \zeta_{2k}, \zeta_{2k})(1 - F_1(t - \zeta_{2k}))(1 - F_2(\zeta_{2k}))dt = \\
\int_{t_{2k} - \zeta_{2k}}^{t_{2k+1} - \zeta_{2k}} v_1(x, \zeta_{2k})(1 - F_1(x))(1 - F_2(\zeta_{2k}))dx = \\
\int_{\zeta_{2k-1}}^{\zeta_{2k+1}} v_1(x, \zeta_{2k})(1 - F_1(x))(1 - F_2(\zeta_{2k}))dx.
\end{aligned} \tag{13}$$

Similarly, for the interval $[t_{2k+1}, t_{2k+2}]$ we have

$$\begin{aligned}
\int_{t_{2k+1}}^{t_{2k+2}} v(\sigma_1, \sigma_2)(1 - F_1(\sigma_1))(1 - F_2(\sigma_2))dt = \\
\int_{t_{2k+1}}^{t_{2k+2}} v_2(\zeta_{2k+1}, t - \zeta_{2k+1})(1 - F_1(\zeta_{2k+1}))(1 - F_2(t - \zeta_{2k+1}))dt = \\
\int_{t_{2k+1} - \zeta_{2k+1}}^{t_{2k+2} - \zeta_{2k+1}} v_2(\zeta_{2k+1}, x)(1 - F_1(\zeta_{2k+1}))(1 - F_2(x))dx = \\
\int_{\zeta_{2k}}^{\zeta_{2k+2}} v_2(\zeta_{2k+1}, x)(1 - F_1(\zeta_{2k+1}))(1 - F_2(x))dx.
\end{aligned} \tag{14}$$

Substituting (13) and (14) into (12), we obtain a new form for the minimization problem:

$$E_u(\zeta_1, \ldots, \zeta_n) =$$
$$\sum_{k=0}^{\infty} \Bigg[ \left(1 - F_2(\zeta_{2k})\right) \int_{\zeta_{2k-1}}^{\zeta_{2k+1}} v_1(x, \zeta_{2k})(1 - F_1(x)) dx +$$
$$\left(1 - F_1(\zeta_{2k+1})\right) \int_{\zeta_{2k}}^{\zeta_{2k+2}} v_2(\zeta_{2k+1}, x)(1 - F_2(x)) dx \Bigg] \to \min \tag{15}$$

(for the sake of generality, we assume $\zeta_{-1} = 0$).

The minimization problem (15) is equivalent to the original problem (4), and the dependency between their solutions is described by (9) and (10). The only constraint for the new problem follows from the fact that the processes are alternating for non-negative periods of time:

$$\begin{cases} \zeta_0 = 0 < \zeta_2 \leq \ldots \leq \zeta_{2n} \leq \ldots \\ \zeta_1 < \zeta_3 \leq \ldots \leq \zeta_{2n+1} \leq \ldots \end{cases} \tag{16}$$

The expression (15) reaches its optimal values either when

$$\frac{dE_u}{d\zeta_k} = 0 \text{ for } k = 1, \ldots, n, \ldots, \tag{17}$$

or on the border described by (16). However, for two processes we can, without loss of generality, ignore the border case. Indeed, assume that $\zeta_i = \zeta_{i+2}$ for some $i > 1$ (one of the processes skips its turn). We can construct a new schedule by removing $\zeta_{i+1}$ and $\zeta_{i+2}$:

$$\zeta_1, \ldots, \zeta_{i-1}, \zeta_i, \zeta_{i+3}, \zeta_{i+4}, \zeta_{i+5}, \ldots$$

It is easy to see that the process described by this schedule is exactly the same process as described by the original one, but the singularity point has been removed.

Thus, at each step the time spent by the processes is determined by (17). We can see that $\zeta_{2k}$ appears in three subsequent terms of $E_u(\sigma_1, \ldots, \sigma_n)$:

$$\ldots + (1 - F_1(\zeta_{2k-1})) \int_{\zeta_{2k-2}}^{\zeta_{2k}} v_2(\zeta_{2k-1}, x)(1 - F_2(x)) dx +$$

$$(1 - F_2(\zeta_{2k})) \int_{\zeta_{2k-1}}^{\zeta_{2k+1}} v_1(x, \zeta_{2k})(1 - F_1(x)) dx +$$

$$(1 - F_1(\zeta_{2k+1})) \int_{\zeta_{2k}}^{\zeta_{2k+2}} v_2(\zeta_{2k+1}, x)(1 - F_2(x)) dx + \ldots.$$

Differentiating (15) by $\zeta_{2k}$, therefore, yields

$$\frac{dE_u}{d\zeta_{2k}} = v_2(\zeta_{2k-1}, \zeta_{2k})(1 - F_1(\zeta_{2k-1}))(1 - F_2(\zeta_{2k})) -$$

$$f_2(\zeta_{2k}) \int_{\zeta_{2k-1}}^{\zeta_{2k+1}} v_1(x, \zeta_{2k})(1 - F_1(x))dx +$$

$$(1 - F_2(\zeta_{2k})) \int_{\zeta_{2k-1}}^{\zeta_{2k+1}} \frac{\partial v_1}{\partial t_2}(x, \zeta_{2k})(1 - F_1(x))dx -$$

$$v_2(\zeta_{2k+1}, \zeta_{2k})(1 - F_1(\zeta_{2k+1}))(1 - F_2(\zeta_{2k})) =$$
$$(1 - F_2(\zeta_{2k}))(v_2(\zeta_{2k-1}, \zeta_{2k})(1 - F_1(\zeta_{2k-1})) - v_2(\zeta_{2k+1}, \zeta_{2k})(1 - F_1(\zeta_{2k+1})) -$$

$$f_2(\zeta_{2k}) \int_{\zeta_{2k-1}}^{\zeta_{2k+1}} v_1(x, \zeta_{2k})(1 - F_1(x))dx +$$

$$(1 - F_2(\zeta_{2k})) \int_{\zeta_{2k-1}}^{\zeta_{2k+1}} \frac{\partial v_1}{\partial t_2}(x, \zeta_{2k})(1 - F_1(x))dx.$$

A similar expression can be derived by differentiating (15) by $\zeta_{2k+1}$. Combining these expressions with (17) gives us the following theorem:

**Theorem 1 (The chain theorem for two processes)**
*The value for $\zeta_{i+1}$ for $i \geq 2$ can be computed for given $\zeta_{i-1}$ and $\zeta_i$ using the formulas*

$$\frac{f_2(\zeta_{2k})}{1 - F_2(\zeta_{2k})} = \frac{v_2(\zeta_{2k-1}, \zeta_{2k})(1 - F_1(\zeta_{2k-1})) - v_2(\zeta_{2k+1}, \zeta_{2k})(1 - F_1(\zeta_{2k+1}))}{\int_{\zeta_{2k-1}}^{\zeta_{2k+1}} v_1(x, \zeta_{2k})(1 - F_1(x))dx} +$$
$$\frac{\int_{\zeta_{2k-1}}^{\zeta_{2k+1}} \frac{\partial v_1}{\partial t_2}(x, \zeta_{2k})(1 - F_1(x))dx}{\int_{\zeta_{2k-1}}^{\zeta_{2k+1}} v_1(x, \zeta_{2k})(1 - F_1(x))dx}, \quad i = 2k+1, \tag{18}$$

$$\frac{f_1(\zeta_{2k+1})}{1 - F_1(\zeta_{2k+1})} = \frac{v_1(\zeta_{2k}, \zeta_{2k+1})(1 - F_2(\zeta_{2k})) - v_1(\zeta_{2k+2}, \zeta_{2k+1})(1 - F_2(\zeta_{2k+2}))}{\int_{\zeta_{2k}}^{\zeta_{2k+2}} v_2(\zeta_{2k+1}, x)(1 - F_2(x))dx} +$$
$$\frac{\int_{\zeta_{2k}}^{\zeta_{2k+2}} \frac{\partial v_2}{\partial t_1}(\zeta_{2k+1}, x)(1 - F_2(x))dx}{\int_{\zeta_{2k}}^{\zeta_{2k+2}} v_2(\zeta_{2k+1}, x)(1 - F_2(x))dx}, \quad i = 2k+2. \tag{19}$$

**Corollary 1** *For the linear cost function (5), the value for $\zeta_{i+1}$ for $i \geq 2$ can be computed for given $\zeta_{i-1}$ and $\zeta_i$ using the formulas*

$$\frac{f_2(\zeta_{2k})}{1 - F_2(\zeta_{2k})} = \frac{F_1(\zeta_{2k+1}) - F_1(\zeta_{2k-1})}{\int_{\zeta_{2k-1}}^{\zeta_{2k+1}} (1 - F_1(x))dx}, \; i = 2k+1, \tag{20}$$

$$\frac{f_1(\zeta_{2k+1})}{1 - F_1(\zeta_{2k+1})} = \frac{F_2(\zeta_{2k+2}) - F_2(\zeta_{2k})}{\int_{\zeta_{2k}}^{\zeta_{2k+2}} (1 - F_2(x))dx}, \; i = 2k+2. \tag{21}$$

The proof follows immediately from the fact that $v_i(t_1, t_2) = a + b$.

Theorem 1 allows us to formulate an algorithm for building an optimal solution. This algorithm is presented in the next subsection.

### 4.2 Optimal Solution for Two Processes: an Algorithm

The goal of the scheduling algorithm is to minimize the expression (15)

$$E_u(\zeta_1, \ldots, \zeta_n) =$$

$$\sum_{k=0}^{\infty} \Bigg[ (1 - F_2(\zeta_{2k})) \int_{\zeta_{2k-1}}^{\zeta_{2k+1}} v_1(x, \zeta_{2k})(1 - F_1(x)) dx +$$

$$(1 - F_1(\zeta_{2k+1})) \int_{\zeta_{2k}}^{\zeta_{2k+2}} v_2(\zeta_{2k+1}, x)(1 - F_2(x)) dx \Bigg] \to \min$$

under the constraints

$$\begin{cases} \zeta_0 = 0 < \zeta_2 \leq \ldots \leq \zeta_{2n} \leq \ldots \\ \zeta_1 < \zeta_3 \leq \ldots \leq \zeta_{2n+1} \leq \ldots. \end{cases}$$

Assume that $A_1$ acts first ($\zeta_1 > 0$). From Theorem 1 we can see that the values of $\zeta_0 = 0$ and $\zeta_1$ determine the set of possible values for $\zeta_2$, the values of $\zeta_1$ and $\zeta_2$ determine the possible values for $\zeta_3$, and so on.

Therefore, a non-zero value for $\zeta_1$ provides us with a tree of possible values of $\zeta_k$. The branching factor of this tree is determined by the number of roots of (18) and (19). Each possible sequence $\zeta_1, \zeta_2, \ldots$ can be evaluated using (15).

For the cases where the total time is limited as discussed in Section 4.5, or where the series in that expression converge, e.g., when each process has a finite cost of finding a solution, the algorithm stops after a finite number of points. In some cases, however, such as for extremely heavy-tailed distributions, it is possible that the above series diverge. To ensure a finite number of iterations in such cases, we set an upper limit on the maximal expected cost.

Another limit is added for the probability of failure. Since $t_i = \zeta_{i-1} + \zeta_i$, the probability that both runs would not be able to find a solution after $t_i$ is

$$(1 - F_1(\zeta_{i-1}))(1 - F_2(\zeta_i)).$$

Therefore, if the difference

$$(1 - F_1(\zeta_{i-1}))(1 - F_2(\zeta_i)) - (1 - p_1)(1 - p_2)$$

becomes small enough, we can conclude that both runs failed to find a solution and stop the execution.

For each value of $\zeta_1$ we can find the best sequence using one of the standard search algorithms, such as Branch-and-Bound. Let us denote the value of the best sequence for each $\zeta_1$ by $E_u(\zeta_1)$. Performing global optimization of $E_u(\zeta_1)$ by $\zeta_1$ provides us with an optimal solution for the case where $A_1$ acts first. Note that the value of $\zeta_1$ may also be 0 ($A_2$ acts first), so we need to compare the value obtained by optimization of $\zeta_1$ with the value obtained by optimization of $\zeta_2$ where $\zeta_1 = 0$.

The flow of the algorithm is illustrated in Figure 9, the formal scheme is presented in Figure 10, and the description of the main routine (realized by the DFS Branch and Bound method) in Figure 11.

The algorithm considers two main branches, one for $A_1$ and one for $A_2$, and they are processed by procedure *minimize_sequence_by_first_point* (Figure 10). At each step, we initialize the array of $\zeta$ values, and pass it, through the procedure *build_optimal_sequence*, to the recursive procedure *df_sbnb*, which represents the core of the algorithm (Figure 11).

The *df_sbnb* procedure, shown in Figure 11, acts as follows. It obtains as an input the array of $\zeta$ values, the cost involved up to the current moment, and the best value reached till now. If the cost exceeds this value, the procedure performs a classical Branch-and-Bound cutoff (lines 1-2).

The inner loop (lines 4-19) corresponds to different roots of the expressions (18) and (19). The new value of $\zeta$ corresponding to $\zeta_k$ is calculated by the procedure *calculate_next_zeta* (line 5), and it cannot exceed the previously found root saved in *last_zeta* (for the first iteration, *last_zeta* is initialized to $\zeta_{k-2}$), lines 3 and 8. Lines 6-7 correspond to the case where the lower bound passed to *calculate_next_zeta* exceeds the maximal available time, and in this case the procedure is stopped.

After the new possible value of $\zeta$ is found, the procedure updates the current cost (line 9), and the stopping criteria mentioned above are validated for the new array of $\zeta$ values, which is denoted as a concatenation of the old array and the new value of $\zeta$ (line 10). If the task is accomplished, the cost is verified versus the best known value (which is updated if necessary), and the procedure returns (lines 10-16). Otherwise, $\zeta$ is temporarily added to the array of $\zeta$, and the Branch-and-Bound procedure is called recursively for calculation $\zeta_{k+1}$.

When the whole tree is traversed (except the cutoffs), the best known cost is returned (line 20). The corresponding array of $\zeta$ is the required solution.

Figure 13 shows a trace of a single Branch-and-Bound run for the example shown in Section 2.2 starting with the optimal value of $\zeta_1$. The optimal schedule derived from the the run is 679, 2393, 7815, 17184 with expected cost of 1216.49 steps. The scheduling points are given in subjective times. Using objective (total) time the schedule can be written as 679, 3072, 10208, and 25000. In this particular run there were no Branch-and-Bound cutoffs due to the small number of roots of (18) and (19).

## 4.3 Necessary Conditions for an Optimal Solution for $n$ Processes

In this section we generalize our solution from the case of two processes to the case of $n$ processes.

Assume that we have $n$ processes $A_1, \ldots, A_n$ using shared resources. One of the possible ways to present a schedule is to use a sequence

$$\langle (A_{i_1}, \Delta t_1), (A_{i_2}, \Delta t_2), \ldots, (A_{i_j}, \Delta t_j), \ldots \rangle,$$

where $A_{i_j}$ is the $j$-th active process, and $\Delta t_j$ is the time allocated for this invocation of $A_{i_j}$.

To simplify the formalization of the problem, however, we use the following alternative representation. First, we allow $\Delta t_j$ to be 0, which makes possible it to represent every schedule as

$$\langle (A_1, \Delta t_1), (A_2, \Delta t_2), \ldots, (A_n, \Delta t_n), (A_1, \Delta t_{n+1}), (A_2, \Delta t_{n+2}), \ldots, (A_n, \Delta t_{2n}), \ldots \rangle.$$

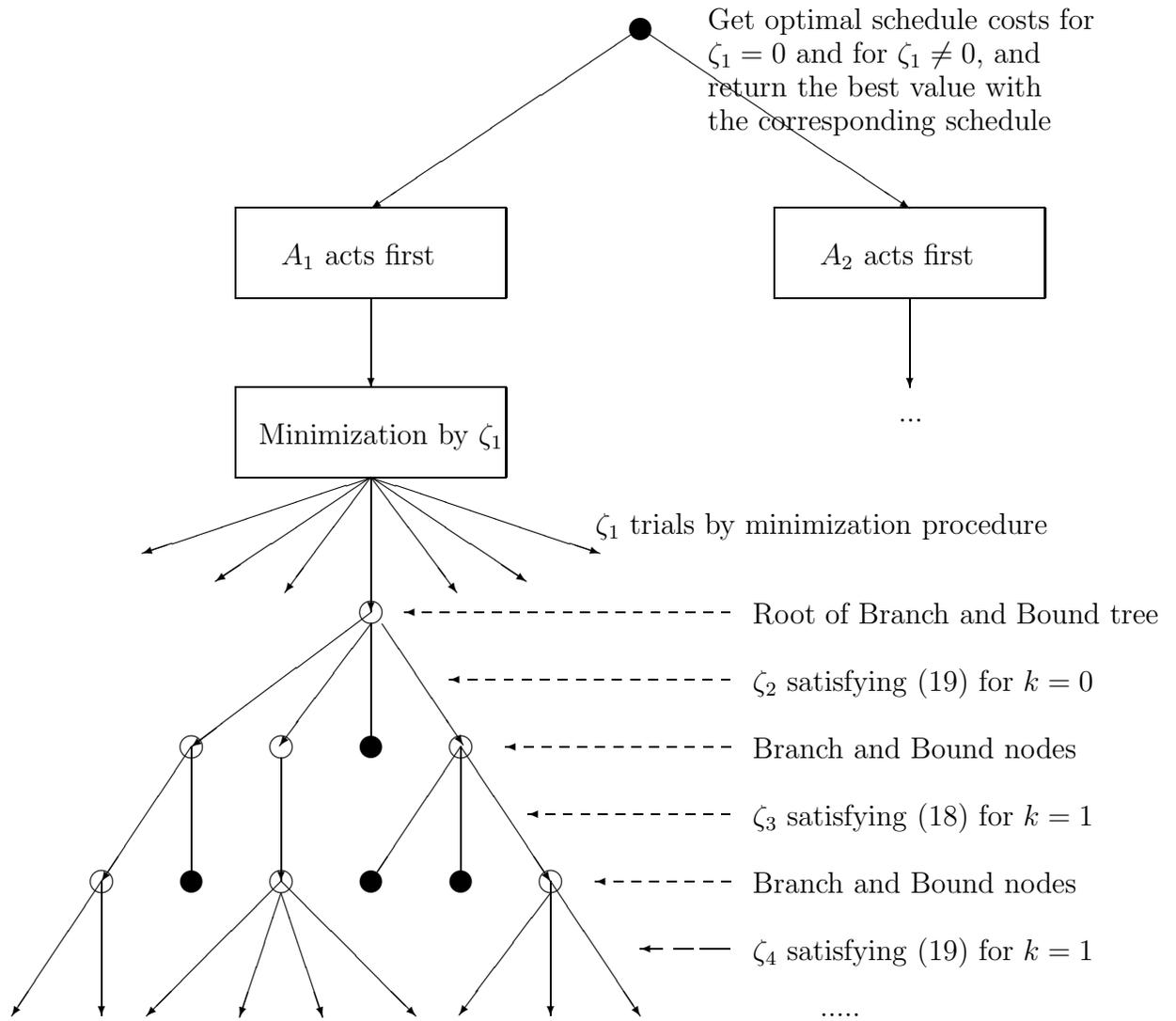

○   Branch and Bound non-leaf nodes

●   Leaf nodes (terminating condition satisfied) and cutoff nodes (expected result is
worse than the already known). The cost is calculated in accordance with (15).

Figure 9: The flow of the algorithm for constructing optimal schedules for 2 processes

---

**procedure** *optimize*

    Input: $F_1(t), F_2(t)$ (performance profiles).

    Output: An optimal sequence and its value.

    $[sequence_1, val_1] \leftarrow minimize\_sequence\_by\_first\_point(A_1)$

    $[sequence_2, val_2] \leftarrow minimize\_sequence\_by\_first\_point(A_2)$

    **if** $val_1 < val_2$ **then**

        **return** $[sequence_1, val_1]$

    **else**

        **return** $[sequence_2, val_2]$

    **end**

**end**

 

**procedure** *minimize\_sequence\_by\_first\_point(process)*

    $zetas[-1] \leftarrow 0$

    $zetas[0] \leftarrow 0$

    **if** $process = A_2$ **then**

        $zetas[1] \leftarrow 0$

    **end**

    Using one of the standard minimization methods, find *zetas*,

    minimizing the value of the function *build\_optimal\_sequence(zetas)*,

    and the corresponding cost.

**end**

Figure 10: Procedure *optimize* builds an optimal sequence for the case when $A_1$ starts, an optimal sequence for the case when $A_2$ starts, compares the results, and returns the best one. Procedure *minimize\_sequence\_by\_first\_point* returns an optimal sequence and its value.

---

**procedure** $build\_optimal\_sequence(zetas)$
    $curr\_cost \leftarrow calculate\_cost(zetas)$
    **return** $dfsbnb(zetas, curr\_cost, MAX\_VALUE)$
**end**

**procedure** $dfsbnb(zetas, curr\_cost, thresh)$
1:    **if** $(curr\_cost \geq thresh)$ **then**
2:        **return** $MAX\_VALUE$     // Cutoff
3:    $last\_value \leftarrow zetas[length(zetas) - 2]$     // The previous time value
4:    **repeat**
5:        $\zeta \leftarrow calculate\_next\_zeta(zetas, last\_value)$
6:        **if** $(\zeta = last\_value)$ **then**     // Skip
7:            **return** $thresh$
8:        $last\_value \leftarrow \zeta$
9:        $delta\_cost \leftarrow calculate\_partial\_cost(zetas, \zeta)$
10:      **if** $(task\_accomplished([zetas \,\|\, \zeta]))$ **then**     // Leaf
11:         **if** $(curr\_cost + delta\_cost < thresh)$ **then**
12:            $optimal\_zetas \leftarrow [zetas \,\|\, \zeta]$
13:            $thresh \leftarrow curr\_cost + delta\_cost$
14:        **end**
15:        **return** $thresh$
16:      **end**
17:      $tmp\_result \leftarrow dfsbnb([zetas \,\|\, \zeta], curr\_cost + delta\_cost, thresh)$
18:      $thresh = min(thresh, tmp\_result)$
19:    **end**     // repeat
20:    **return** $thresh$
**end**

---

Figure 11: Procedure $build\_optimal\_sequence$, given the prefix of the time sequence, restores the optimal sequence with this prefix using the DFS Branch and Bound search algorithm, and returns the sequence itself and its value. $[x \,\|\, y]$ stands for concatenation $x$ and $y$. Auxiliary functions are shown in Figure 12.

1. *calculate_cost(zetas)* computes the cost of the sequence (or its part) in accordance with (15),

2. *calculate_partial_cost(zetas, ζ)* computes the additional cost obtained by adding ζ to the sequence,

3. *calculate_next_zeta(zetas, last_value)* uses (18) or (19) to calculate the value of the next ζ that is greater than *last_value*. If no such a solution exists, the maximal time value is returned,

4. *task_accomplished(zetas)* returns *true* when the task may be considered to be accomplished (e.g., either maximal possible time is over, or the probability of error is negligible, or the upper limit on the cost is exceeded).

Figure 12: Auxiliary functions used in the optimal schedule algorithm

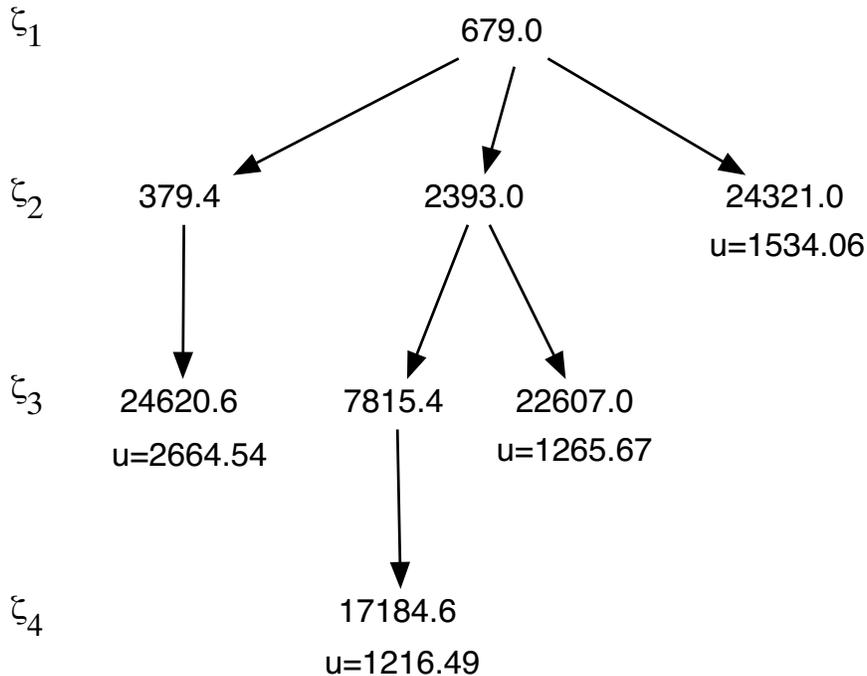

Figure 13: A trace of a single run of the Branch-and-Bound procedure starting with the optimal value of $\zeta_1$.

Therefore, the system alternates between $n$ states $S_1 \to S_2 \to \ldots \to S_n \to S_1 \to \ldots$, where the state $S_i$ corresponds to the situation where $A_i$ is active and the rest of the processes are idle. The time spent in the $k$-th invocation of $S_i$ is $\Delta t_{kn+i}$.

As in the case of two processes, we call the time interval corresponding to the sequence of states $S_1 \to S_2 \to \ldots \to S_n$ a *phase* and denote phase $k$ by $\Phi_k$. We denote the process switch points of $\Phi_k$ by $t_k^1, t_k^2, \ldots, t_k^n$, where

$$t_k^i = \sum_{j=0}^{k-1} \Delta t_{nj+i}.$$

Process $A_i$ is active in phase $k$ in the interval $[t_k^{i-1}, t_k^i]$, and the entire phase lasts from $t_k^0$ to $t_k^n$. The corresponding scheme is shown in Figure 14.

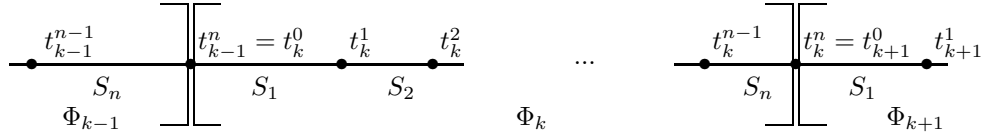

Figure 14: Notations for times, states and phases for $n$ processes

To simplify the following discussion, we would like to allow indices $i$ in $t_k^i$ to be less than 0 or greater than $n$. For this purpose, we denote

$$t_k^i = t_{k+\lfloor i/n \rfloor}^{i \bmod n}, \tag{22}$$

and the index of the process active in the interval $[t_k^{i-1}, t_k^i]$ we denote by $\#i$. For $i \bmod n \neq 0$ we obtain $\#i = i \bmod n$, while for $i \bmod n = 0$ we have $\#i = n$. Notation (22) claims that the shift by $n$ in the upper index is equivalent to the shift by 1 in the phase number:

$$t_k^{i+n} = t_{k+1}^i.$$

As in the case of two processes, we denote by $\zeta_k^i$ the total time that $A_{\#i}$ has been active up to $t_k^i$. $\zeta_k^i$ corresponds to the cumulative time spent in phases 1 to $k$ in state $S_{\#i}$, and there is a one-to-one correspondence between the sequences of $\zeta_k^i$ and $t_k^i$:

$$\zeta_k^i - \zeta_{k-1}^i = t_k^i - t_k^{i-1}, \tag{23}$$

$$\sum_{j=0}^{n-1} \zeta_k^{i-j} = t_k^i \text{ for } i \geq n. \tag{24}$$

The first equation corresponds to the fact that the time between $t_k^{i-1}$ and $t_k^i$ is accumulated into the $\zeta$ values of process $A_{\#i}$, while the second equation claims that at each switch the objective time of the system is equal to the sum of the subjective times of each process. For the sake of uniformity we also denote

$$\zeta_{-1}^1 = \ldots = \zeta_{-1}^n = \zeta_0^0 = 0.$$

By construction of $\zeta_k^i$ we can see, that at time interval $[t_k^{i-1}, t_k^i]$ the subjective time of process $A_j$ has the following form:

$$\sigma_j(t) = \begin{cases} \zeta_k^j, & j = 1, \ldots, i-1, \\ (t - t_k^{i-1}) + \zeta_{k-1}^i, & j = i, \\ \zeta_{k-1}^j, & j = i+1, \ldots, n. \end{cases} \tag{25}$$

The subjective time functions for a system with 3 processes are illustrated in Figure 15.

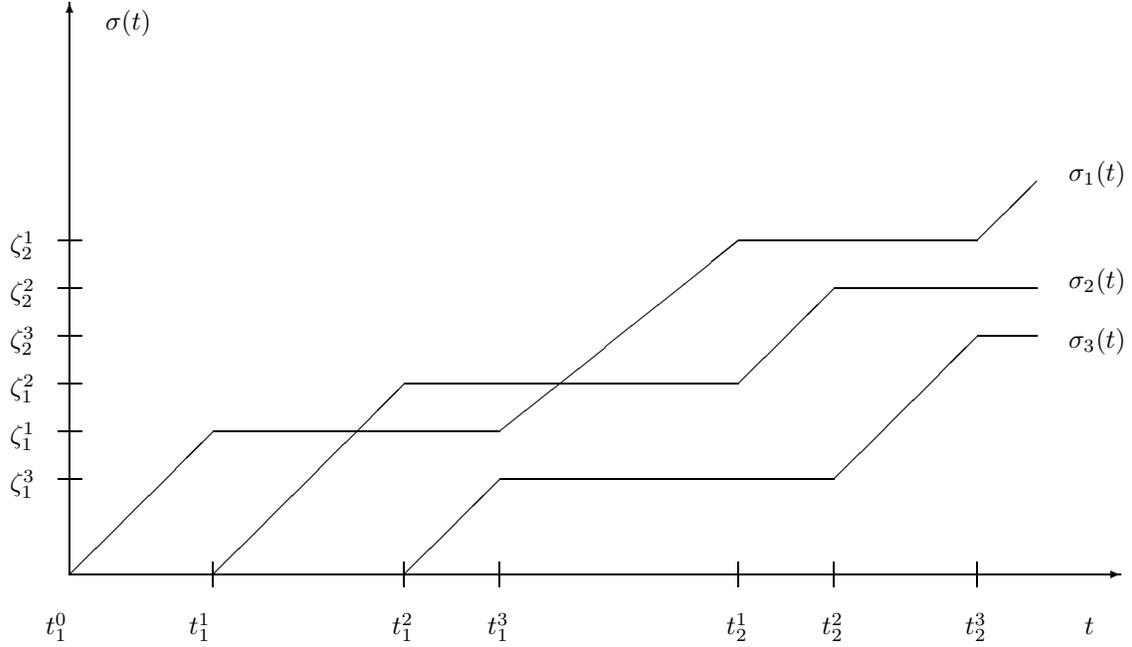

Figure 15: Subjective time functions for a system with 3 processes

To find an optimal schedule for a system with $n$ processes, we need to minimize the expression given by (6). The only constraints are the monotonicity of the sequence of $\zeta$ for each process $i$:

$$\zeta_k^i \le \zeta_{k+1}^i \text{ for each } k, i. \tag{26}$$

Given the expressions for $\sigma_j$, we can prove the following lemma:

**Lemma 1** *For a system of $n$ processes, the expression for the expected cost (6) can be rewritten as*

$$E_u(\zeta_1, \ldots, \zeta_n, \ldots) = \sum_{k=0}^{\infty} \sum_{i=1}^{n} \prod_{j=i+1}^{i+n-1} (1 - F_{\#j}(\zeta_{k-1}^j)) \int_{\zeta_{k-1}^i}^{\zeta_k^i} (1 - F_i(x)) dx. \tag{27}$$

The proof is given in Appendix A.1.

This lemma makes it possible to prove the chain theorem for an arbitrary number of processes:

**Theorem 2 (The chain theorem)** *The value for $\zeta_{m+1}^{l-1}$ may either be $\zeta_m^{l-1}$, or can be computed given the previous $2n-2$ values of $\zeta$ using the formula*

$$\frac{f_l(\zeta_m^l)}{1-F_l(\zeta_m^l)} = \frac{\prod\limits_{j=l+1}^{l+n-1}(1-F_{\#j}(\zeta_{m-1}^j)) - \prod\limits_{j=l+1}^{l+n-1}(1-F_{\#j}(\zeta_m^j))}{\sum\limits_{i=l-n+1}^{l-1}\prod\limits_{\substack{j=i+1\\ \#j\neq l}}^{i+n-1}(1-F_{\#j}(\zeta_m^j))\int_{\zeta_m^i}^{\zeta_{m+1}^i}(1-F_{\#i}(x))dx} \tag{28}$$

The proof of the theorem is given in Appendix A.2.

## 4.4 Optimal Solution for $n$ Processes: an Algorithm

The goal of the presented algorithm is to minimize the expression (27)

$$E_u(\zeta_1,\ldots,\zeta_n,\ldots) = \sum_{k=0}^{\infty}\sum_{i=1}^{n}\prod_{j=i+1}^{i+n-1}(1-F_{\#j}(\zeta_{k-1}^j))\int_{\zeta_{k-1}^i}^{\zeta_k^i}(1-F_i(x))dx$$

under the constraints

$$\zeta_k^i \leq \zeta_{k+1}^i \text{ for each } k,i.$$

As in the case of two processes, assume that $A_1$ acts first. By Theorem 2, given $2n-2$ values of $\zeta$

$$\zeta_1^0, \zeta_1^1, \ldots, \zeta_1^n, \zeta_2^1, \zeta_2^2, \zeta_2^{n-3},$$

we can determine all the possibilities for the value of $\zeta_2^{n-2}$ (either $\zeta_1^{n-2}$ if the process skips its turn, or one of the roots of (28)). Given the values up to $\zeta_2^{n-2}$, we can determine the values for $\zeta_2^{n-1}$, and so on.

The idea of the algorithm is similar to the algorithm for two processes. The first $2n-2$ variables (including $\zeta_1^0 = 0$) determine the tree of possible values for $\zeta$. Optimization over $2n-3$ first variables, therefore, provides us with an optimal schedule (as before, we compare the results for the case where the first $k < n$ variables are 0). The only difference from the case of two processes is that a process may skip its turn. However, we can ignore the case when *all* the processes skip their turn, since we can remove such a loop from the schedule. The scheme of the algorithm is presented in Figure 16, and the description of the main routine (realized by the DFS Branch and Bound method) is presented in Figure 17.

## 4.5 Optimal Solution in the Case of Additional Constraints

Assume now that the problem has additional constraints: the solution time is limited by $T$ and the probability of success of the $i$-th process $p_i$ is not necessarily 1.

It is possible to show that the expressions for the distribution function and the expected cost have almost the same form as in the regular framework:

**Claim 2** *Let the system solution time be limited by $T$, and let $p_i$ be the probability of success for the $i$-th process. Then the expressions for the goal-time distribution and expected cost*

---

*Procedure optimize builds n optimal schedules (each process may start first), compares the results, and returns the best one*

**procedure** *optimize*
    $best\_val \leftarrow MAX\_VALUE$
    $best\_sequence \leftarrow \emptyset$
    **loop for** $i$ **from** 1 **to** $n$ **do**
        $[sequence, val] \leftarrow minimize\_sequence\_by\_first\_points(i)$
        **if** $(val < best\_val)$ **then**
            $best\_val \leftarrow val$
            $best\_sequence \leftarrow sequence$
    **end**
    **return** $[best\_sequence, best\_val]$
**end**

// Procedure $minimize\_sequence\_by\_first\_points$ gets as a parameter
// the index of a process which starts, and returns an optimal
// sequence and its value
**procedure** $minimize\_sequence\_by\_first\_points(process\_to\_start)$
    **loop for** $i$ **from** 0 **to** $n - 1$
        $zetas[-i] \leftarrow 0$
    **end**
    **loop for** $i$ **from** 1 **to** $process\_to\_start - 1$
        $zetas[i] \leftarrow 0$
    **end**
    Using one of the standard minimization methods, find $zetas$,
    minimizing the value of the function $build\_optimal\_sequence(zetas)$.
**end**

---

Figure 16: An algorithm for finding an optimal schedule for $n$ processes. The result contains the vector of $\zeta_i$, such that $\zeta_i = \zeta_0^i = \zeta_{\lfloor i/n \rfloor}^{i \bmod n}$.

---

**procedure** *build_optimal_sequence*(*zetas*)
    *curr_cost* ← *calculate_cost*(*zetas*)
    **return** *dfsbnb*(*zetas*, *curr_cost*, *MAX_VALUE*, 0)
**end**

**procedure** *dfsbnb*(*zetas*, *curr_cost*, *thresh*, *nskip*)
    **if** (*curr_cost* ≥ *thresh*) **then**
        **return** *MAX_VALUE*     // Cutoff
    *last_value* ← *zetas*[*length*(*zetas*) − *n*]    // The previous time value for the current process
    **repeat**
        *ζ* ← *calculate_next_zeta*(*zetas*, *last_value*)
        **if** (*ζ* = *last_value*) **then**     // Skip
            **break loop**
        *last_value* ← *ζ*
        *delta_cost* ← *calculate_partial_cost*(*zetas*, *ζ*)
        **if** (*task_accomplished*([*zetas* || *ζ*])) **then**     // Leaf
            **if** (*curr_cost* + *delta_cost* < *thresh*) **then**
                *optimal_zetas* ← [*zetas* || *ζ*]
                *thresh* ← *curr_cost* + *delta_cost*
            **end**
            **break loop**
        **end**
        *tmp_result* ← *dfsbnb*([*zetas* || *ζ*], *curr_cost* + *delta_cost*, *thresh*, 0)
        *thresh* = *min*(*thresh*, *tmp_result*)
    **end**     // repeat
    **if** (*nskip* < *n* − 1) **then**     // Skip is possible
        *zeta* ← *zetas*[*length*(*zetas*) − *n*]
        *tmp_result* ← *dfsbnb*([*zetas* || *ζ*], *curr_cost*, *thresh*, *nskip* + 1)
        *thresh* = *min*(*thresh*, *tmp_result*)
    **end**
    **return** *thresh*
**end**

---

Figure 17: Procedure *build_optimal_sequence*, given the prefix of time sequence, restores the optimal sequence with this prefix using the DFS Branch and Bound search algorithm, and returns the sequence itself and its value. [*x* || *y*] stands for concatenation *x* and *y*. The auxiliary functions used are similar to their counterparts in Figure 12, but deal with *n* processes instead of 2.

*are as follows:*

$$F_n(t, \sigma_1, \ldots, \sigma_n) = 1 - \prod_{i=1}^{n}(1 - p_i F_i(\sigma_i)) \ (for \ t \le T), \tag{29}$$

$$E_u(\sigma_1, \ldots, \sigma_n) = \int_0^T \left( u_t' + \sum_{i=1}^{n} \sigma_i' u_{\sigma_i}' \right) \prod_{i=1}^{n}(1 - p_i F_i(\sigma_i))dt. \tag{30}$$

The proof is similar to the proof of Claim 1.

This claim shows that all the formulas used in the previous sections are valid for the current settings, with three differences:

1. We use $p_j f_j$ instead of $F_j$ and $p_j f_j$ instead of $f_j$.

2. All the integrals are from 0 to $T$ instead of from 0 to $\infty$.

3. All time variables are limited by $T$.

The first two conditions may be easily incorporated into all the algorithms. The last condition implies additional changes in the chain theorems and the algorithms. The chain theorem for $n$ processes now becomes:

**Theorem 3** *The value for $\zeta_k^j$ can either be $\zeta_{k-1}^j$, or it can be computed given the previous $2n - 2$ values of $\zeta$ using formula (28), or it can be calculated by the formula*

$$\zeta_k^j = T - \sum_{l=1}^{n-1} \zeta_k^{j-l}. \tag{31}$$

The first two alternatives are similar to Theorem 2, while the third one corresponds to the boundary condition given by Equation (24). This third alternative adds one more branch to the DFS Branch and Bound algorithm; the rest of the algorithm remains unchanged.

Similar changes in the algorithms are performed in the case of the maximal allowed time $T_i$ per process. In practice, we always use this limitation, setting $T_i$ such that the probability for $A_i$ to reach the goal after $T_i$, $p_i(1 - F_i(T_i))$, becomes negligible.

## 5. Process Scheduling by Intensity Control

In this section we analyze the problem of optimal scheduling for the case of intensity control, which is equivalent to replacing the binary scheduling functions $\theta_i(t)$ with continuous functions with a range between 0 and 1. In this paper we assume a linear cost function of the form (5). We believe, however, that similar analysis is applicable to the setup with any differentiable $u$.

It is easy to see that all the formulas for the distribution function and the expected cost from Claim 1 are still valid under intensity control settings.

For the linear cost function (5), the minimization problem has the form

$$E_u(\sigma_1, \ldots, \sigma_n) = \int_0^\infty \left( a + b \sum_{i=1}^{n} \sigma_i' \right) \prod_{j=1}^{n}(1 - F_j(\sigma_j))dt \to \min. \tag{32}$$

Without loss of generality, we can assume $a + b = 1$. This leads to the equivalent minimization problem

$$E_u(\sigma_1, \ldots, \sigma_n) = \int_0^\infty \left( (1-c) + c \sum_{i=1}^n \sigma_i' \right) \prod_{j=1}^n (1 - F_j(\sigma_j)) dt \to \min, \tag{33}$$

where $c = b/(a + b)$ can be viewed as a normalized resource weight. The constraints, however, are more complicated than for the suspend/resume model:

1. As before, $\sigma_i$ must be continuous, and $\sigma_i(0) = \sigma_i'(0) = 0$ (at the beginning all the processes are idle).

2. We assume $\sigma_i$ to have a partially-continuous derivative $\sigma_i'$, and this derivative should lie between 0 and 1. This requirement follows from the definition of intensity and the fact that $\sigma_i' = \theta_i$: no process can work for a negative amount of time, and no process can work with the intensity greater than the one allowed. Since we consider a framework with shared resources, and the total intensity is limited, we have an additional constraint: the sum of all the derivatives $\sigma_i'$ at any time point cannot exceed 1.

Thus, this optimization problem has the following boundary conditions:

$$\begin{aligned}
&\sigma_i(0) = 0, \ \sigma_i'(0) = 0 \text{ for } i = 1, \ldots, n, \\
&0 \le \sigma_i' \le 1 \text{ for } i = 1, \ldots, n, \\
&0 \le \sum_{i=1}^n \sigma_i' \le 1.
\end{aligned} \tag{34}$$

We are looking for a set of functions $\{\sigma_i\}$ that provide a solution to minimization problem (33) under constraints (34).

Let $g(t, \sigma_1, \ldots, \sigma_n, \sigma_1', \ldots, \sigma_n')$ be a function under the integral sign of (33):

$$g(t, \sigma_1, \ldots, \sigma_n, \sigma_1', \ldots, \sigma_n') = \left( (1-c) + c \sum_{i=1}^n \sigma_i' \right) \prod_{j=1}^n (1 - F_j(\sigma_j)). \tag{35}$$

A traditional method for solving problems of this type is to use the Euler-Lagrange necessary conditions: a set of functions $\sigma_1, \ldots, \sigma_n$ provides a weak (local) minimum to the functional

$$E_u(\sigma_1, \ldots, \sigma_n) = \int_0^\infty g(t, \sigma_1, \ldots, \sigma_n, \sigma_1', \ldots, \sigma_n') dt$$

only if $\sigma_1, \ldots, \sigma_n$ satisfy a system of equations of the form

$$g_{\sigma_k}' - \frac{d}{dt} g_{\sigma_k'}' = 0. \tag{36}$$

We can prove the following lemma:

**Lemma 2** *The Euler-Lagrange conditions for minimization problem (33) yield two strong invariants:*

1. For processes $k_1$ and $k_2$ for which $\sigma_{k_1}$ and $\sigma_{k_2}$ are not on the border described by (34), the distribution and density functions satisfy

$$\frac{f_{k_1}(\sigma_{k_1})}{1 - F_{k_1}(\sigma_{k_1})} = \frac{f_{k_2}(\sigma_{k_2})}{1 - F_{k_2}(\sigma_{k_2})}. \tag{37}$$

2. If the schedules of all the processes are not on the border described by (34), then either $c = 1$ or $f_k(\sigma_k) = 0$ for each $k$.

The proof of the lemma is given in Appendix A.3. The above lemma provides necessary conditions for a local minimum in the inner points described by constraints (34). These conditions, however, are very restricting. Therefore, we look for more general conditions, suitable for boundary points as well[7].

We start with the following lemma:

**Lemma 3** If an optimal solution for minimization problem (33) under constraints (34) exists, then there exists an optimal solution $\sigma_1, \ldots, \sigma_n$, such that at each time $t$ all the resources are consumed, i.e.,

$$\forall t \sum_{i=1}^{n} \sigma_i'(t) = 1. \tag{38}$$

In the case where time cost is not zero ($c \neq 1$), the equality above is a necessary condition for solution optimality.

The proof of the lemma is given in Appendix A.4.

**Corollary 2** Under intensity control settings, as in the case of suspend-resume settings, minimization problem (33) has the form (6), i.e.

$$E_u(\sigma_1, \ldots, \sigma_n) = \int_0^\infty \prod_{j=1}^{n} (1 - F_j(\sigma_j)) dt \to \min.$$

Lemma 3 corresponds to our intuition: if a resource is available, it should be used. Without loss of generality, we restrict our discussion to schedules satisfying (38), even in the case where time cost is zero. This leads to the following invariant:

$$\forall t \sum_{i=1}^{n} \sigma_i(t) = t. \tag{39}$$

Assume now that we have two $F$-equivalent processes $A_1$ and $A_2$ with density function $f(t)$ satisfying the normal distribution law with mean value $m$. Let $t_1$ and $t_2$ be the cumulative time consumed by each of the processes at time $t$, i.e., $\sigma_1(t) = t_1$ and $\sigma_2(t) = t_2$. The question is, which process should be active at $t$ (or should they be active in parallel with partial intensities)?

---

7. Note also that even if the conditions above hold, they do not necessarily provide the optimal solution. Moreover, problems in variation calculus do not necessarily have a minimum, since there is no analogue for the Weierstrass theorem for continuous functions on a closed set.

Without loss of generality, $t_1 < t_2$, which means that the first process is required to cover a larger area to succeed: $1 - F(t_1) > 1 - F(t_2)$. This supports a policy that at time $t$ activates the second process. This policy is further supported if $A_1$ has a lower distribution density, $f_1(t_1) < f_2(t_2)$, as illustrated in Figure 18(a). If, however, the first process has a higher density, as illustrated in Figure 18(b), it is not clear which of the two processes should be activated at time $t$. What is the optimal policy in the general case[8]? The answer relies

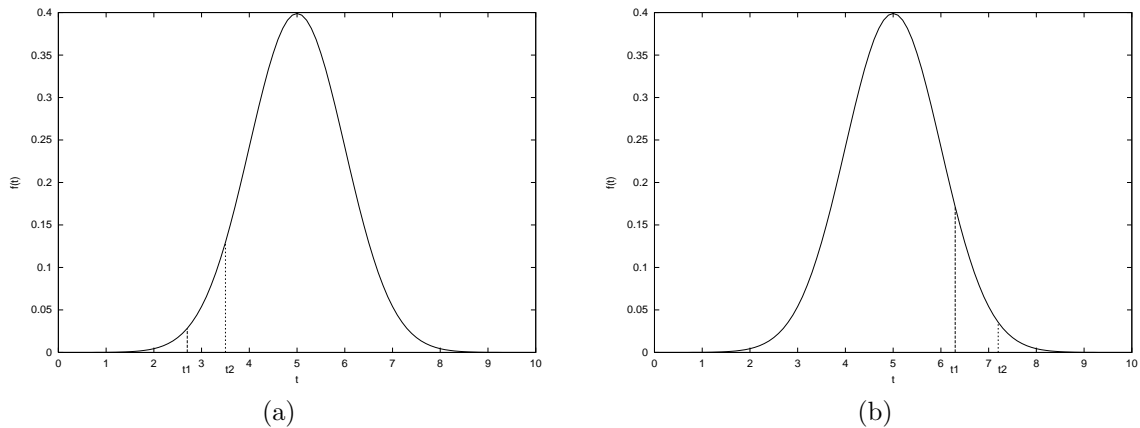

(a)                             (b)

Figure 18: (a) Process $A_1$ (currently at $t_1$) has lower density and larger area to cover, and therefore is inferior. (b) Process $A_1$ has lower density, but smaller area to cover, and the decision is unclear.

heavily on the functions that appear in (37). These functions, described by the equation

$$h_k(t) = \frac{f_k(t)}{1 - F_k(t)}, \tag{40}$$

are known as *hazard functions*, and they play a very important role in the following theorem describing necessary conditions for optimal schedules.

**Theorem 4** *Let the set of functions $\{\sigma_i\}$ be a solution of minimization problem (6) under constraints (34). Let $t_0$ be a point where the hazard functions of all the processes $h_i(\sigma_i(t))$ are continuous, and let $A_k$ be the process active at $t_0$ ($\sigma'_k(t_0) > 0$), such that for any other process $A_i$*

$$h_i(\sigma_i(t_0)) < h_k(\sigma_k(t_0)). \tag{41}$$

*Then at $t_0$ process $k$ consumes all the resources, i.e. $\sigma'_k(t_0) = 1$.*

The proof of the theorem is given in Appendix A.5.

By Theorem 4 and Equation (37), intensity control may only be useful when hazard functions of at least two processes are equal. However, even in this case the equilibrium is not always stable. Assume that within some interval $[t', t'']$ processes $A_i$ and $A_j$ are working with partial intensity, which implies $h_i(\sigma_i(t)) = h_j(\sigma_j(t))$. Assume now that both

---

$h_i(t)$ and $h_j(t)$ are monotonically increasing. If at some moment $t$ we give a priority to one of the processes, it will obtain a higher value of the hazard function, and will get all the subsequent resources. The only case of stable equilibrium is when $h_i(\sigma_i(t))$ and $h_j(\sigma_j(t))$ are monotonically decreasing functions or constants.

The intuitive discussion above is formulated in the following theorem:

**Theorem 5** *An active process will remain active and consume all resources as long as its hazard function is monotonically increasing.*

The proof is given in Appendix A.6.

This theorem imply the important corollary:

**Corollary 3** *If the hazard function of one of the processes is greater than or equal to that of the others at $t = 0$ and is monotonically increasing by $t$, this process should be the only one to be activated.*

We can conclude that the extension of the suspend-resume model to intensity control in many cases does not increase the power of the model and is beneficial only for monotonically decreasing hazard functions. If no time cost is taken into account ($c = 1$), however, the intensity control permits us to connect the two concepts: that of the model with shared resources and that of the model with independent agents:

**Theorem 6** *If no time cost is taken into account ($c = 1$), the model with shared resources under intensity control settings is equivalent to the model with independent processes under suspend-resume control settings. Namely, given a suspend-resume solution for the model with independent processes, we may reconstruct an intensity-based solution with the same cost for the model with shared resources and vice versa.*

The proof of the theorem is given in Appendix A.7.

Theorem 4 claims that if the process with the maximal value of $h_k(\sigma_k(t))$ is active, it will take all the resources. Why, then, would we not always choose the process with the highest value of $h_k(\sigma_k(t))$ to be active? It turns out that such a strategy is not optimal. Let us consider two processes with the distribution densities shown in Figure 19(a). The corresponding values of the hazard functions are shown in Figure 19(b). If we were using the above strategy, $A_2$ would be the only active process. Indeed, at time $t = 0$, $h_2(\sigma_2(0)) > h_1(\sigma_1(0))$, which would lead to the activation of $A_2$. After that moment, $A_1$ would remain idle and its hazard function remain 0. This strategy would result in an expected time of 2. If, on the other hand, we would have activated $A_1$ only, the result would be an expected time of 1.5. Thus, although $h_1(\sigma_1(0)) < h_2(\sigma_2(0))$, it is better to give all the resources to $A_1$ from the beginning due to its superiority in the future.

A more elaborate example is shown in Figure 20. It corresponds to the case of two processes that are not $F$-equivalent, one of which is a linear combination of two normal distributions, $f(t) = 0.5f_{N(0.6,0.2)}(t) + 0.5f_{N(4.0,2.0)}(t)$, where $f_{N(\mu,\sigma)}(t)$ is the distribution density of normal distribution with mean value $\mu$ and standard deviation $\sigma$, and the second process is uniformly distributed in $[1.5, 2.5]$. Activating $A_1$ only results in $0.5 \times 0.6 + 0.5 \times 4.0 = 2.3$, activating $A_2$ only results in an expected time of 2.0, while activating $A_1$ for time 1.2 followed by activating $A_2$ results in (approximately) $0.6 \times 0.5 + (1.2 + 2.0) \times 0.5 = 1.9$.

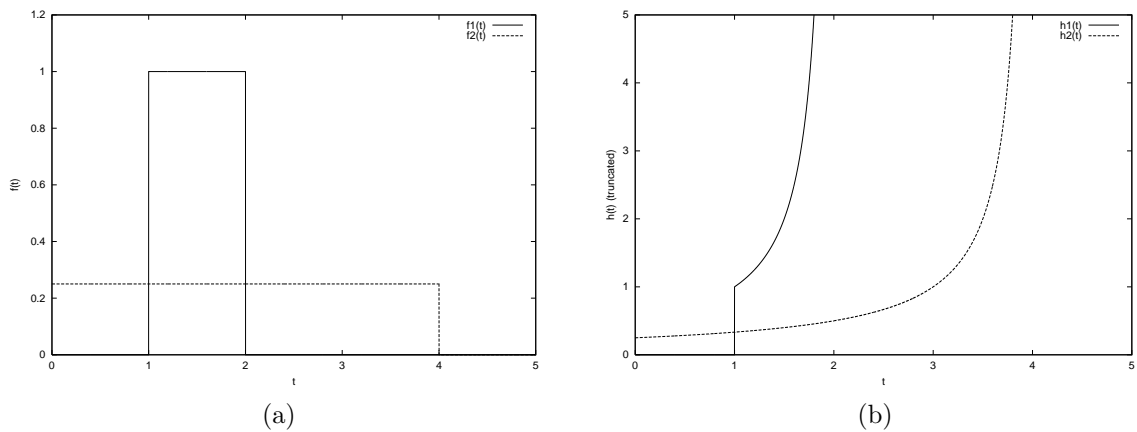

Figure 19: The density function and the hazard function for two processes. Although $h_1(\sigma_1(0)) < h_2(\sigma_2(0))$, it is better to give all the resources to $A_1$.

The best solution is, therefore, to start the execution by activating $A_1$, and at some point $t'$ transfer the control to $A_2$. In this case we interrupt an active process with a greater value of hazard function, preferring an idle process with a zero value of hazard function (since $h_1(\sigma_1(t')) > h_2(\sigma_2(t')) = 0$).

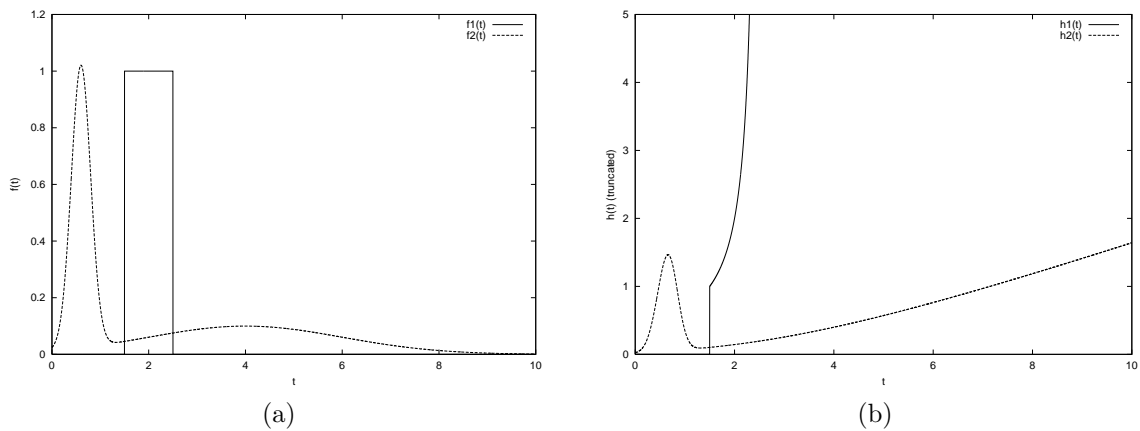

Figure 20: The density function and the hazard function for two processes. The best solution is to start with $A_1$, and at some point interrupt it in favor of $A_2$, although the latter has a zero hazard function.

These examples show that a straightforward use of hazard functions for building optimal schedules can be very problematic. However, since the suspend-resume model is a specific case of the intensity control model, the hazard functions still may be useful for understanding the behavior of optimal schedules, and this is used in the next section.

## 6. Optimal Scheduling for Standard Distributions

In this section we present the results of the optimal scheduling strategy for a system of processes whose performance profiles meet one of the well-known distributions: uniform, exponential, normal and lognormal. Then we show the results for processes with bimodal and multimodal distribution functions.

We have implemented three scheduling policies for two agents:

1. **Sequential strategy**, which schedules the processes one after another, initiating the second process when the probability that the first one will find a solution becomes negligible. For processes that are not $F$-equivalent, we choose the best order of process invocation.

2. **Simultaneous strategy**, which simulates a simultaneous execution of both processes.

3. **Optimal strategy**, which is an implementation of the algorithm described in Section 4.2.

In the rest of this section we compare these three strategies, when no deadline is given, and the processes are stopped when the probability that they can still find a solution becomes negligible.

Our goal is to compare different scheduling strategies and not to analyze the behavior of the processes. Absolute quantitative measurements, such as average cost, are very process dependent, and therefore are not appropriate for scheduling strategy evaluation. We therefore would like to normalize the results of the application of different scheduling methods to minimize the effect of the process behavior. In the case of $F$-equivalent processes, a good candidate for the normalization coefficient is the expected time of the individual process. For processes that are not $F$-equivalent, however, the decision is not straightforward, and therefore we use the results of the sequential strategy as the normalization factor.

We define the *relative quality* $q_{ref}(S)$ of strategy $S$ with respect to strategy $S_{ref}$ as

$$q_{ref}(S) = 1 - \frac{\bar{u}(S)}{\bar{u}(S_{ref})}, \tag{42}$$

where $\bar{u}(S)$ is the average cost of strategy $S$. This measurement corresponds to the gain (maybe negative) of strategy $S$ relative to the reference strategy. In this section we use the sequential strategy as our reference strategy.

### 6.1 Uniform Distribution

Assume that the goal-time distribution of the processes meets the uniform law over the interval $[t_0, T]$, i.e., has distribution functions

$$F(t) = \begin{cases} 0 & \text{if } t < t_0, \\ (t - t_0)/(T - t_0) & \text{if } t \in [t_0, T], \\ 1 & \text{if } t > T \end{cases} \tag{43}$$

and density functions

$$f(t) = \begin{cases} 0 & \text{if } t \notin [t_0, T], \\ 1/(T - t_0) & \text{if } t \in [t_0, T]. \end{cases} \tag{44}$$

The density function of a process uniformly distributed in $[0, 1]$ is shown in Figure 21(a).

The hazard function of the uniform distribution has the form

$$h(t) = \begin{cases} 0 & \text{if } t < t_0, \\ \dfrac{1/(T - t_0)}{1 - (t - t_0)/(T - t_0)} = \dfrac{1}{T - t} & \text{if } t \in [t_0, T], \end{cases} \tag{45}$$

which is a monotonically increasing function. By Corollary 3, only one process will be active, and the optimal strategy should be equivalent to the sequential strategy. If the processes are not $F$-equivalent, the problem can be solved by choosing the process with the minimal expected time.

A more interesting setup involves a uniformly distributed process that is not guaranteed to find a solution. This case corresponds to a probability of success $p$ that is less than 1. As it was claimed in Section 4.5, the corresponding distribution and density function should be multiplied by $p$. As a result, the hazard function becomes

$$h(t) = \begin{cases} 0 & \text{if } t < t_0, \\ \dfrac{p}{(T - t_0) - p(t - t_0)} & \text{if } t \in [t_0, T]. \end{cases} \tag{46}$$

This function is still monotonically increasing by $t$, and the conclusions remain the same. The graphs for hazard functions of processes uniformly distributed in $[0, 1]$ with probability of success of 0.5, 0.8 and 1 are shown in Figure 21(b).

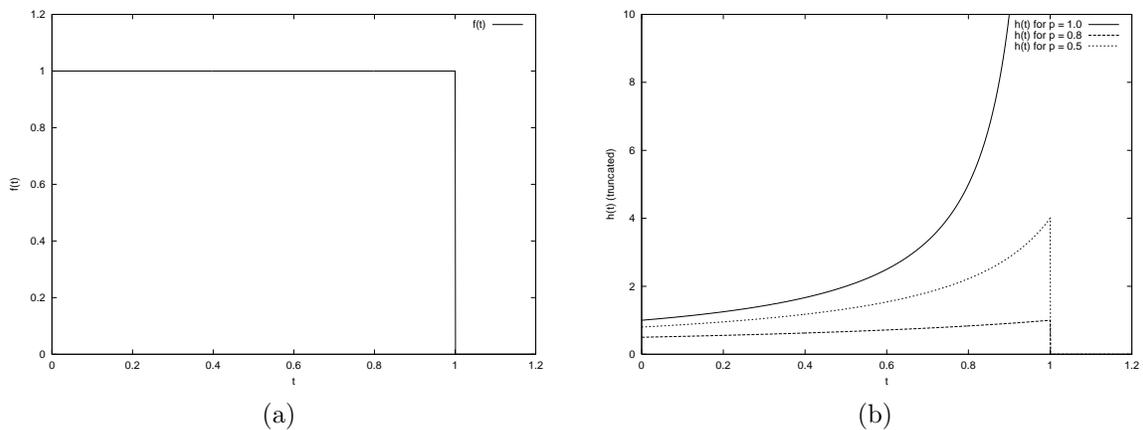

Figure 21: (a) The density function of a process, uniformly distributed in $[0, 1]$, (b) hazard functions for processes uniformly distributed in $[0, 1]$ with probability of success of 0.5, 0.8 and 1.

## 6.2 Exponential Distribution

The exponential distribution is described by the density function

$$f(t) = \begin{cases} 0 & \text{if } t \leq 0 \\ \lambda e^{-\lambda t} & \text{if } t > 0, \end{cases} \tag{47}$$

and the distribution function has the form

$$F(t) = \begin{cases} 0 & \text{if } t \le 0 \\ 1 - e^{-\lambda t} & \text{if } t > 0. \end{cases} \tag{48}$$

Substituting these expressions into (6) gives

$$E_u(\sigma_1, \ldots, \sigma_n) = \int_0^\infty \prod_{j=1}^n (1 - F_j(\sigma_j)) dt = \int_0^\infty e^{-\sum_{j=1}^n \lambda_j \sigma_j(t)} dt.$$

For a system with $F$-equivalent processes, by Lemma 3

$$\sum_{j=1}^n \lambda_j \sigma_j(t) = \lambda \sum_{j=1}^n \sigma_j(t) = \lambda t,$$

and therefore

$$E_u(\sigma_1, \ldots, \sigma_n) = \int_0^\infty e^{-\lambda t} dt = \frac{1}{\lambda}.$$

Thus, for a system with $F$-equivalent processes all the schedules are equivalent. This interesting fact is reflected also in the behavior of the hazard function, which is constant:

$$h(t) \equiv \lambda.$$

However, if the probability of success is smaller than 1, the hazard function becomes a monotonically decreasing function:

$$h(t) = \frac{p \lambda e^{-\lambda t}}{1 - p(1 - e^{-\lambda t})} = \frac{p \lambda}{p + (1 - p) e^{\lambda t}}.$$

Such processes should work simultaneously (with identical intensities for $F$-equivalent processes, and with intensities maintaining the equilibrium of hazard functions otherwise), since each process which has been idle for a while has an advantage over its working teammate.

Figure 22(a) shows the density function of an exponentially distributed process with $\lambda = 1$. The graphs for the hazard functions of processes exponentially distributed with $\lambda = 1$ and probability of success of 0.5, 0.8 and 1 are shown in Figure 22(b).

Let us consider a somewhat more elaborate example, involving processes that are not $F$-equivalent. Assume that we have two learning systems, both with an exponential-like performance profile typical of such systems. We also assume that one of the systems requires a delay for preprocessing but works faster. Thus, we assume that the first system has a distribution density $f_1(t) = \lambda_1 e^{-\lambda_1 t}$, and the second one has a density $f_2(t) = \lambda_2 e^{-\lambda_2 (t - t_2)}$, such that $\lambda_1 < \lambda_2$ (the second is faster), and $t_2 > 0$ (it also has a delay). Assume that both learning systems are deterministic over a given set of examples, and that they may fail to learn the concept with the same probability of $1 - p = 0.5$. The graphs for the density and hazard functions of the two systems are shown in Figure 23.

We applied the optimal scheduling algorithm of Section 4.2 for the values $\lambda_1 = 3$, $\lambda_2 = 10$, and $t_2 = 5$. The optimal schedule is to activate the first system for 1.15136 time units, then (if it found no solution) to activate the second system for 5.77652 time units.

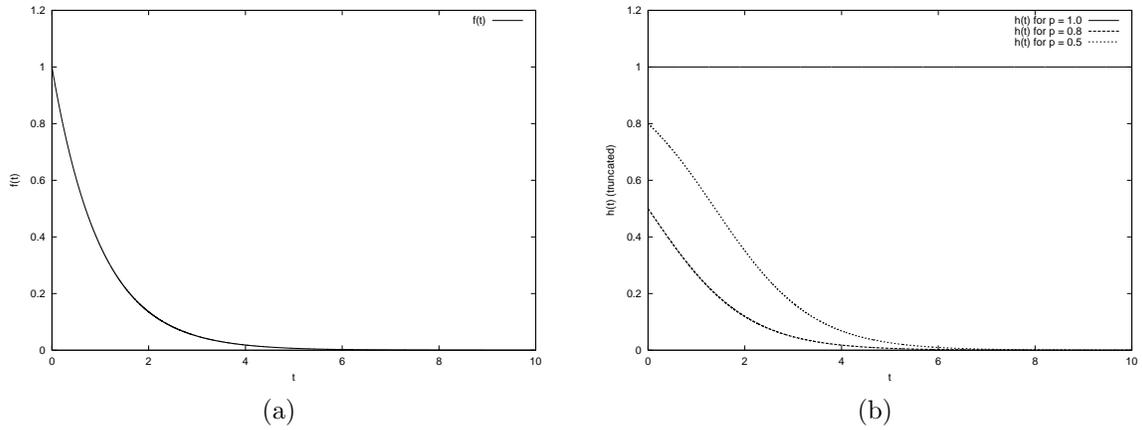

(a)                                             (b)

Figure 22: (a) The density function of a process, exponentially distributed with $\lambda = 1$, (b) hazard functions for processes exponentially distributed with $\lambda = 1$ and probability of success of 0.5, 0.8 and 1.

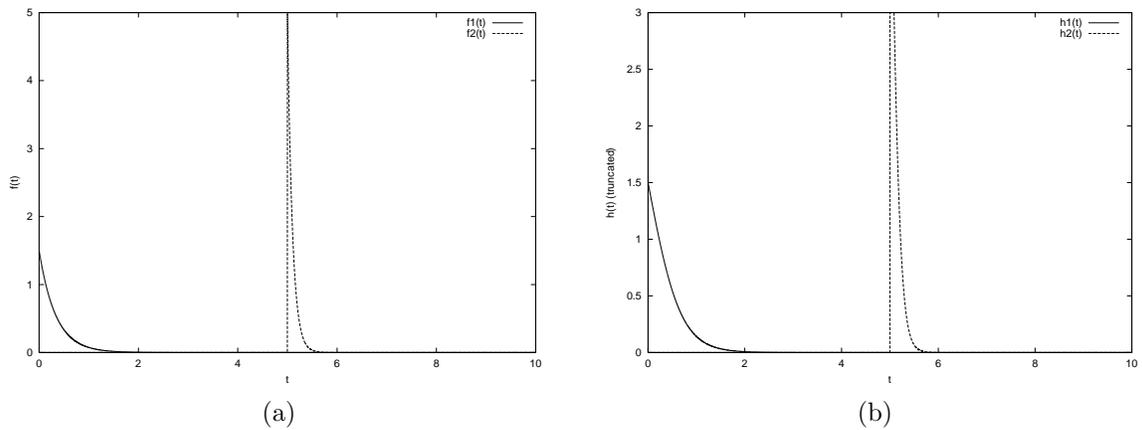

(a)                                             (b)

Figure 23: (a) Density and (b) hazard functions for two exponentially distributed systems, with different values of $\lambda$ and time shift.

Then the first system will run for additional 3.22276 time units, and finally the second system will run for 0.53572 time units. If at this point no solution has been found, both systems have failed with a probability of $1 - 10^{-6}$ each.

Figure 24(a) shows the relative quality of the simultaneous and optimal scheduling strategies as a function of $t_2$ for $p = 0.8$ (for 10000 simulated examples). For large values of $t_2$ the benefit of switching from the first algorithm to the second decreases, and this is reflected in the relative quality of the optimal strategy. The simultaneous strategy, as we can see, is beneficial only for relatively small values of $t_2$.

Figure 24(b) reflects the behavior of the strategies for a fixed value of $t_2 = 5.0$ as a function of probability of success $p$. The simultaneous strategy is inferior, and its quality decreases while $p$ increases. Indeed, when the probability of success is 1, running the second algorithm and the first one simultaneously will be a waste of time. On the other hand, the optimal strategy has a positive benefit, which means that the resulting schedules are not trivial.

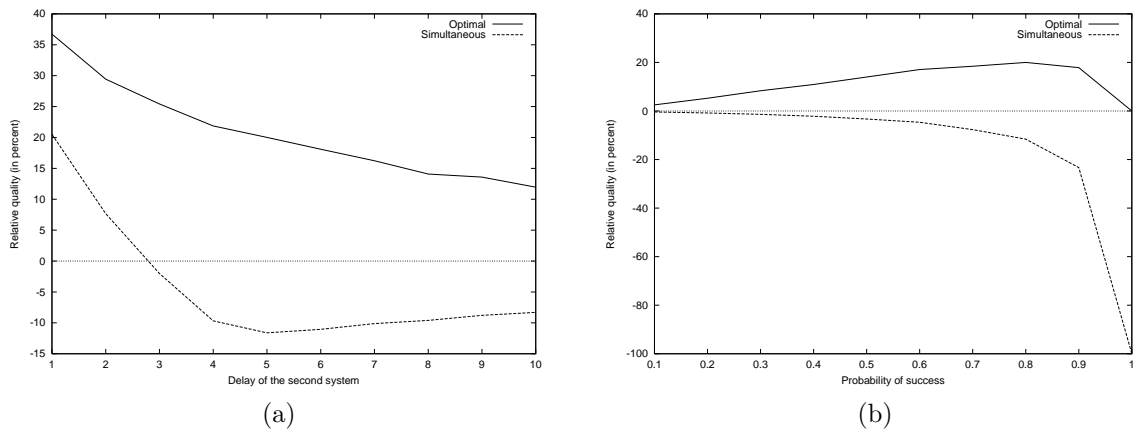

(a)                                (b)

Figure 24: Learning systems: Relative quality of optimal and simultaneous scheduling strategies (a) as a function of $t_2$ for fixed $p = 0.8$, and (b) as a function of $p$ for fixed $t_2 = 5$.

## 6.3 Normal Distribution

The normal distribution with mean value $m$ and deviation $\sigma$ is described by the density function

$$f(t) = \frac{1}{\sqrt{2\pi}\sigma} e^{-\frac{(t-m)^2}{2\sigma^2}}, \tag{49}$$

and its distribution function is

$$F(t) = \frac{1}{\sqrt{2\pi}\sigma} \int_{-\infty}^{t} e^{-\frac{(x-m)^2}{2\sigma^2}} dx. \tag{50}$$

Since we use $t_0 = 0$, we should have used a truncated normal distribution with a distribution density

$$\frac{1}{(1-\mu)} \cdot \frac{1}{\sqrt{2\pi}\sigma} e^{-\frac{(t-m)^2}{2\sigma^2}},$$

and a distribution function

$$\frac{1}{1-\mu} \cdot \left[ \frac{1}{\sqrt{2\pi}\sigma} \int_{-\infty}^{t} e^{-\frac{(x-m)^2}{2\sigma^2}} \, dx - \mu \right],$$

where

$$\mu = \frac{1}{\sqrt{2\pi}\sigma} \int_{-\infty}^{0} e^{-\frac{(x-m)^2}{2\sigma^2}} \, dx,$$

but if $m$ is large enough, $\mu$ may be considered to be 0. The density function of a normally distributed process with $m = 5$ and $\sigma = 1$ is shown in Figure 25(a).

The hazard function of a normal distribution is monotonically increasing, which leads to the same conclusions as for a uniform distribution. However, a probability of success of less than 1 completely changes the behavior of the hazard function: after some point, it starts to decrease. The graphs for hazard functions of processes normally distributed with a mean value of 5, standard deviation of 1 and probabilities of success of 0.5, 0.8 and 1 are shown in Figure 25(b).

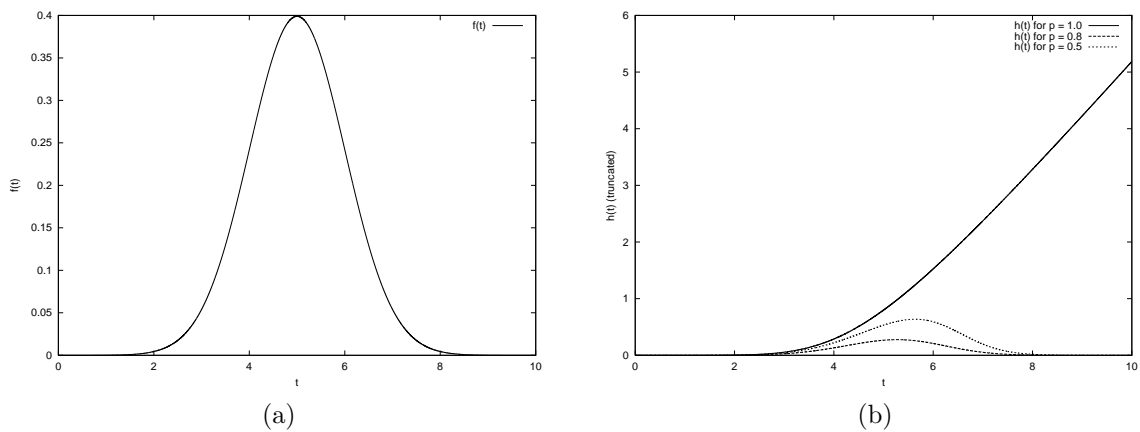

(a)                                                      (b)

Figure 25: (a) The density function of a normally distributed process, with $m = 5$ and $\sigma = 1$, (b) hazard functions for normally distributed processes with $m = 5$ and $\sigma = 1$, with the probabilities of success of 0.5, 0.8 and 1.

As in the previous example, we now consider a case of two processes that are not $F$-equivalent, running with the same deviation $\sigma = 1$ and the same probability of success $p$. The first process is assumed to have $m_1 = 1$, while the second process is started with some delay $\Delta m$. The relative quality for 10000 simulated examples is shown in Figure 26. Figure 26(a) shows the relative quality as a function of $\Delta m$ for $p = 0.8$; Figure 26(b) shows the relative quality as a function of $p$ for $\Delta m = 2$. Unlike exponential distribution, the gain for this example for the optimal strategy is rather small.

## 6.4 Lognormal Distribution

The random variable $X$ is lognormally distributed, if $\ln X$ is normally distributed. The density function and the distribution function with the corresponding parameters $m$ and $\sigma$

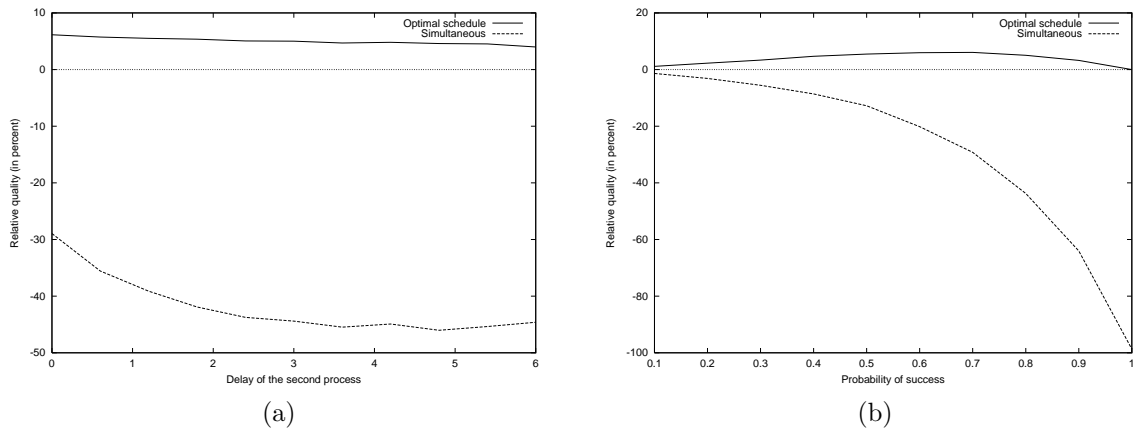

(a)                                        (b)

Figure 26: Normal distribution: relative quality (a) as a function of $\Delta m$ for fixed $p = 0.8$, and (b) as a function of $p$ for fixed $\Delta m = 2$.

can be written as

$$f(t) = \frac{1}{t\sqrt{2\pi}\sigma} e^{-\frac{(\log(t)-m)^2}{2\sigma^2}}, \tag{51}$$

$$F(t) = \frac{1}{\sqrt{2\pi}\sigma} \int_{-\infty}^{\log(t)} e^{-\frac{(x-m)^2}{2\sigma^2}} dx. \tag{52}$$

Lognormal distribution plays a significant role in AI applications since in many cases search time is distributed under the lognormal law. The density function of the lognormal distribution with mean value of $\log(5.0)$ and standard deviation of $1.0$ is shown in Figure 27(a), and the hazard functions for different values of $p$ are shown in Figure 27(b). Let us consider a simulated experiment similar to its analogue for normal distribution. We consider two processes that are not $F$-equivalent, with the parameters $\sigma = 1$ and the same probability of success $p$. The first process is assumed to have $m_1 = 1$, while the second process is started with some delay, such that $m_2 - m_1 = \Delta m > 0$. The relative quality for 10000 simulated examples is shown in Figure 28. Figure 28(a) shows the relative quality as a function of $\Delta m$ for $p = 0.8$; Figure 28(b) shows the relative quality as a function of $p$ for $\Delta m = 2$. The graphs show that for small values of $\Delta m$ both the optimal and the simultaneous strategy have a significant benefit over the sequential one. However, for larger values, the performance of the optimal strategy approaches the performance of the sequential strategy, while the simultaneous strategy becomes inferior.

## 6.5 Bimodal and Multimodal Density Functions

Experiments show that in the case of $F$-equivalent processes with a unimodal distribution function, the sequential strategy is often optimal. In this section we consider less trivial distributions.

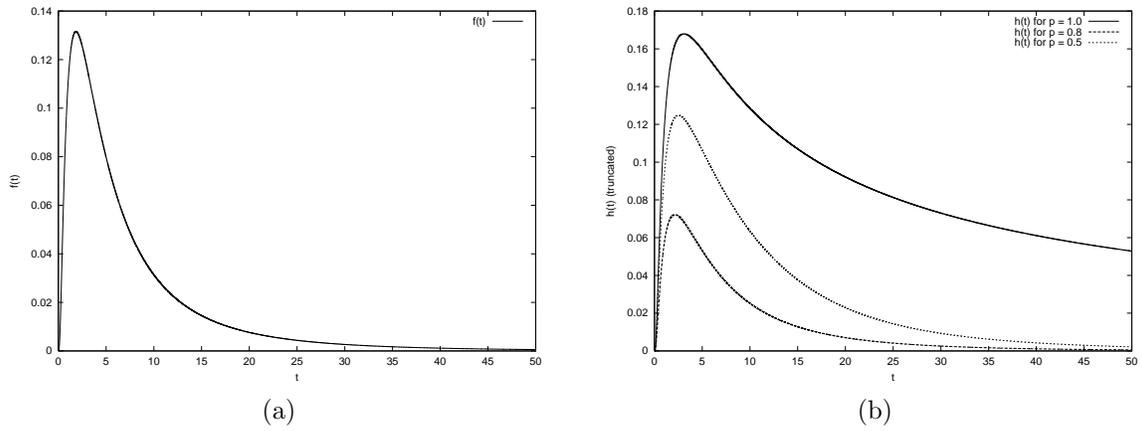

(a)

(b)

Figure 27: (a) Density function for lognormal distribution with mean value of log(5.0) and standard deviation of 1.0 and (b) hazard functions for lognormally distributed processes with mean value of log(5.0), standard deviation of 1, and the probabilities of success of 0.5, 0.8 and 1.

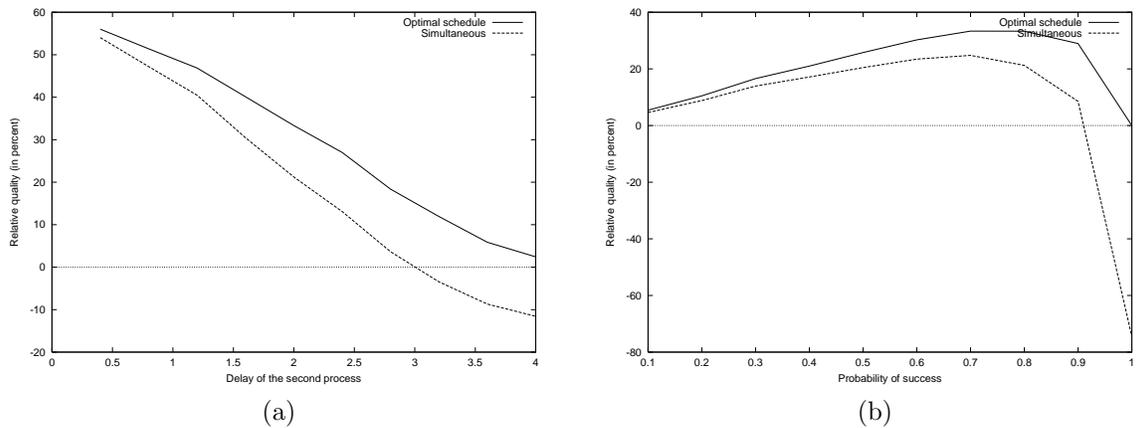

(a)

(b)

Figure 28: Lognormal distribution: relative quality (a) as a function of $\Delta m$ for fixed $p = 0.8$, and (b) as a function of $p$ for fixed $\Delta m = 2$.

Assume first that we have a non-deterministic algorithm with a performance profile expressed by a linear combination of two normal distributions with the same deviation:

$$f(t) = 0.5 f_{N(\mu_1, \sigma)} + 0.5 f_{N(\mu_2, \sigma)}.$$

An example of the density and hazard functions of such distributions with $\mu_1 = 2, \mu_2 = 5$, and $\sigma = 0.5$ is given in Figure 29.

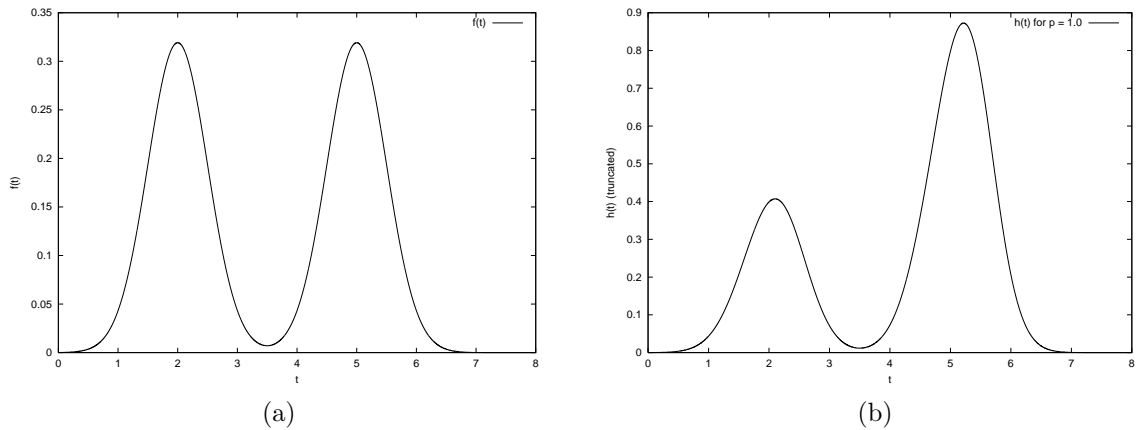

(a)                               (b)

Figure 29: (a) Density function and (b) hazard function for a process distributed according to the density function $f(t) = 0.5 f_{N(2, 0.5)} + 0.5 f_{N(5, 0.5)}$ with the probability of success of $p = 0.8$.

Assume that we invoke two runs of this algorithm with fixed values of $\mu_1 = 2$, $\sigma = 0.5$, and $p = 0.8$, and the free variable $\mu_2$. Figure 30 shows how the relative quality of the scheduling strategies is influenced by the distance between the peaks, $\mu_2 - \mu_1$. The results correspond to the intuitive claim that the larger distance between the peaks, the more attractive the optimal and the simultaneous strategies become.

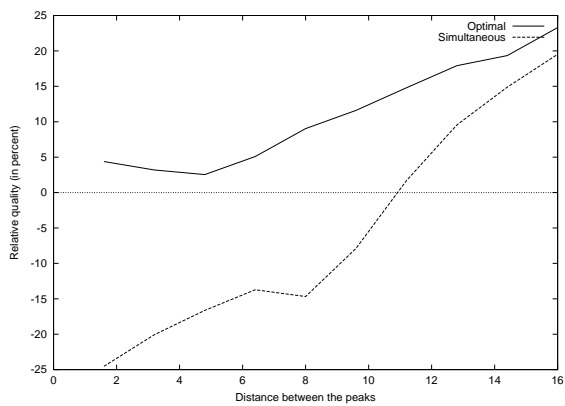

Figure 30: Bimodal distribution: relative quality as a function of the distance between the peaks.

Now let us see how the number of peaks of the density function affects the scheduling quality. We consider a case of partial uniform distribution, where the density is distributed over $k$ identical peaks of length 1 placed symmetrically in the time interval from 0 to 100. (Thus, the density function will be equal to $1/k$ when $t$ belongs to one of such peaks, and 0 otherwise.) In this experiment we have chosen $p = 1$.

Figure 31 shows the relative quality of the system as a function of $k$, obtained for 10000 randomly generated examples. We can see from the results, that the simultaneous strategy is inferior, due to the "valleys" in the distribution function. The optimal strategy returns schedules where the processes switch after each peak, but the relative quality of the schedules decreases as the number of peaks increases.

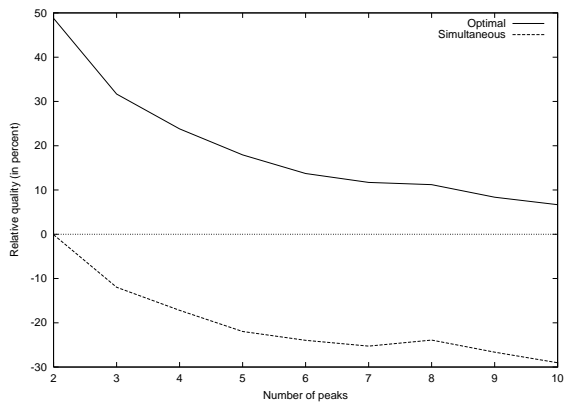

Figure 31: Multimodal distribution: relative quality as a function of the number of peaks.

## 7. Experiments: Using Optimal Scheduling for the Latin Square Problem

To test the performance of our algorithm in a realistic domain, we applied it to the Latin Square problem described in Section 2.2. We assume that we are given a Latin Square problem with two initial configurations, and a fully deterministic algorithm with distribution function and distribution density shown in Figure 7.

We compare the performance of the schedule produced by our algorithm to the performance of the sequential and simultaneous strategies described in Section 6. In addition, we test a schedule which runs the processes one after another, allowing a single switch at the *optimal* point (an analogue of the restart technique for two processes). We refer to this schedule as a single-point restart schedule.

Note that the case of two initial configurations corresponds to the case of two processes in our framework. In general, we could think of a *set* of $n$ initial configurations that would correspond to $n$ processes. For sufficiently large $n$, the restart strategy where each restart starts with a different initial configuration, becomes close to optimal.

Our experiments were performed for different values of $N$, with 10% of the square pre-colored. The performance profile was induced based on a run of 50,000 instances, and the remaining 50,000 instances were used as 25,000 testing pairs. All the schedules were applied with a fixed deadline $T$, which corresponds to the maximal allowed number of generated nodes.

Since the results of the sequential strategy in this type of problems are much worse than the results of other strategies for sufficiently large values of $T$, we instead used the simultaneous strategy as the reference in the relative quality measure.

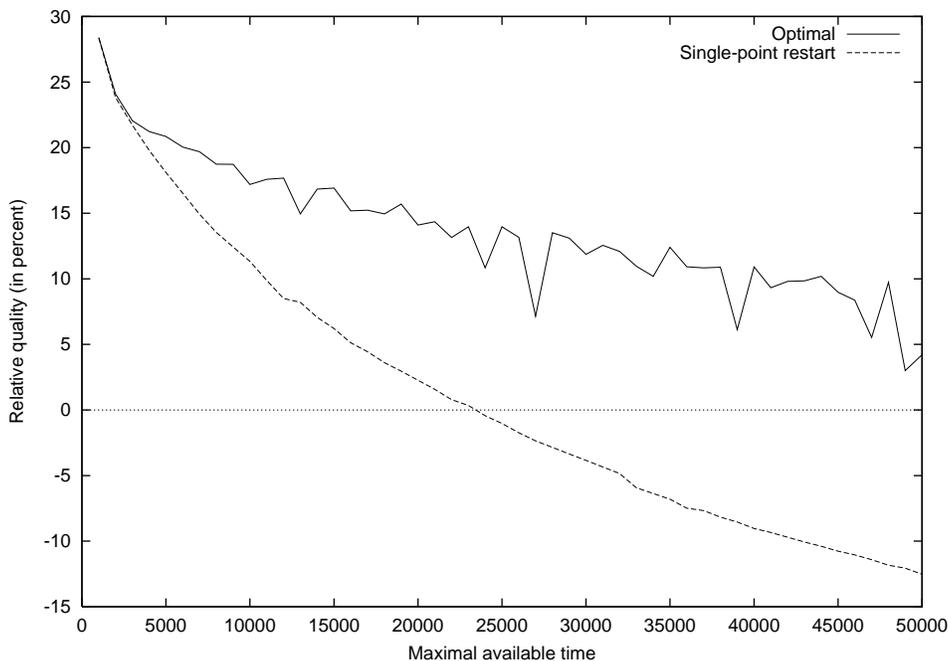

Figure 32: Relative quality as a function of maximal allowed time $T$

Figure 32 shows how maximal available time $T$ (the $x$ axis) influences the quality of the schedules (the $y$ axis), where the simultaneous strategy has been used as a reference.

For small values of $T$, both single-point restart and the optimal strategy have about a 25% gain over the simultaneous strategy, since they produce schedules which are close to the sequential one. However, when available time $T$ increases, the benefit of parallelization becomes more significant, and the simultaneous strategy overcomes the single-point restart strategy. The relative quality of the optimal schedule also decreases when $T$ increases, since the resulting schedule contains more switches between the two problem instances being solved.

Figure 33 illustrates how the optimal and single-point restart schedules relate to the simultaneous schedule for different size Latin Squares (given $T = 25,000$). The initial gain of both strategies is about 50%. However, for the problems with $N = 20$ the single-point restart strategy becomes worse than the simultaneous one. For larger sizes the probability of solving the Latin Square problem with a time limit of $25,000$ steps becomes smaller and smaller, and the benefit of the optimal strategy also approaches zero.

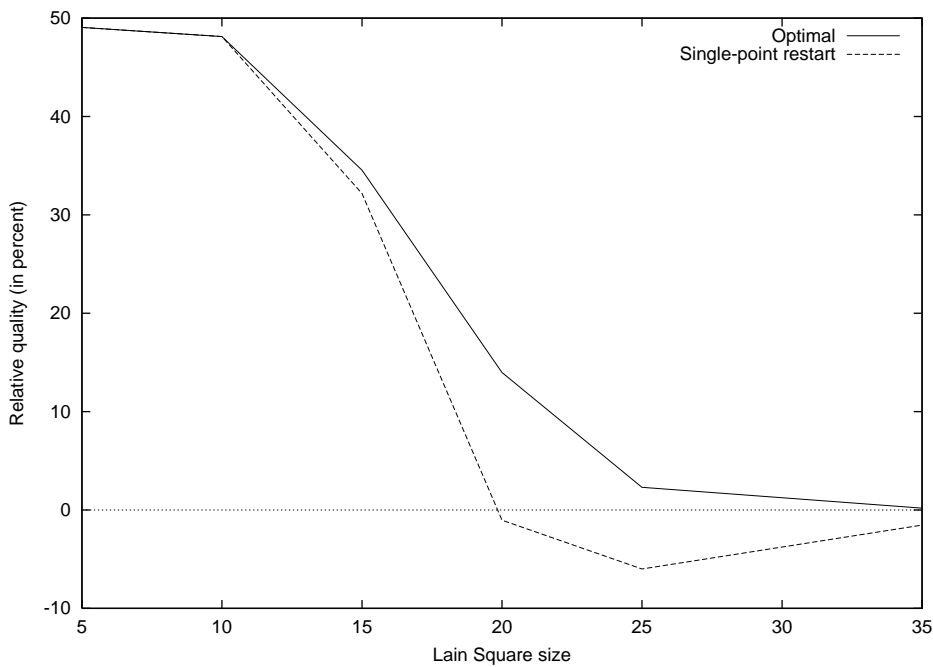

Figure 33: Relative quality as a function of the size of the Latin Square

# 8. Combining Restart and Scheduling Policies

Luby, Sinclair, and Zuckerman (1993) showed that the restart strategy is optimal if an infinite number of identical runs are available. When this number is limited, the restart strategy is not optimal. Sometimes, however, we have a mixed situation. Assume that we have two initial states, a *non-deterministic* algorithm, and a linear time cost. On one hand, we can perform restarts of a run corresponding to one of the initial states. On the other hand, we can switch between the runs corresponding to the two initial states. What would be an optimal policy in this case?

The expected time of a run based on a single initial state is

$$E(t^*) = \frac{1}{F(t^*)} \int_0^{t^*} (1 - F(t)) dt, \tag{53}$$

where $t^*$ is the restart point and $F(t)$ is the distribution function. This formula is obtained by a simple summation of the geometric series with coefficient $1 - F(t^*)$, and is a continuous form of the formula given by Luby, Sinclair, and Zuckerman (1993). Minimization of (53) by $t^*$ gives us the optimal restart point.

Assume first that the sequence of restarts on a single initial state is a process interruptible only at the restart points. Since the probability of failure of $i$ successive restarts is $(1 - F(t^*))^i$, this process is exponentially distributed. Thus, the problem is reduced to scheduling of two exponentially distributed processes. According to the analysis in Section 6.2, all schedules are equivalent if the problems corresponding to the two initial states are solvable. Otherwise, the optimal policy is to alternate between the two processes at each restart point.

A more interesting case is when we allow rescheduling at any time point. In general, it is not beneficial to switch between the processes in non-restart points (otherwise these rescheduling points would have been chosen for restart). Such rescheduling, however, can be beneficial if the cost associated with restarts is higher than the rescheduling cost[9].

Let us assume that each restart has a constant cost $C$. Similarly to (53), we can write the expected cost of a policy performing restarts at point $t^{**}$ as

$$E(t^{**}) = \frac{1}{F(t^{**})} \int_0^{t^{**}} (1 - F(t))dt + \frac{1 - F(t^{**})}{F(t^{**})^2} C, \qquad (54)$$

where the second term corresponds to the series

$$0 + C(1 - F(t^{**})) + 2C(1 - F(t^{**}))^2 + \dots$$

Let $t^{**}$ and $t^*$ be the optimal restart points for the setups with and without associated costs respectively. $t^{**}$ should be greater than $t^*$ due to the restart cost.

Let us consider the following schedule: the first process runs for $t^*$, then the second process runs for $t^*$, then the first process runs (with no restart) for additional $t^{**} - t^*$, then the second process runs for additional $t^{**} - t^*$. Then the first process restarts and runs for $t^*$ and so forth.

Let us compare the expected time of such schedule with the time of the pure restart policy, where the first process runs for $t^{**}$, then the second process runs for $t^{**}$, then the first process restarts and runs for $t^{**}$ and so forth.

Similarly to (15), the expected time of the first schedule in the interval $[0, 2t^{**}]$ can be written as

$$E_{sched} = \int_0^{t^*} (1 - F(t))dt + (1 - F(t^*)) \int_0^{t^*} (1 - F(t))dt +$$

$$(1 - F(t^*)) \int_{t^*}^{t^{**}} (1 - F(t))dt + (1 - F(t^{**})) \int_{t^*}^{t^{**}} (1 - F(t))dt.$$

On the other hand, the expected time of the second schedule in the same interval is

$$E_{simple} = \int_0^{t^{**}} (1 - F(t))dt + (1 - F(t^{**})) \int_0^{t^{**}} (1 - F(t))dt =$$

$$(2 - F(t^{**})) \int_0^{t^{**}} (1 - F(t))dt.$$

---

9. An example for such setup is robotic search, where returning the robot to the initial position is more expensive than suspending and resuming the robot.

$E_{sched}$ can be rewritten as

$$E_{sched} = \int_0^{t^*} (1 - F(t))dt + (1 - F(t^*)) \int_0^{t^{**}} (1 - F(t))dt +$$

$$(1 - F(t^{**})) \int_0^{t^{**}} (1 - F(t))dt - (1 - F(t^{**})) \int_0^{t^*} (1 - F(t))dt =$$

$$F(t^{**}) \int_0^{t^*} (1 - F(t))dt + (2 - F(t^*) - F(t^{**})) \int_0^{t^{**}} (1 - F(t))dt =$$

$$F(t^{**}) \int_0^{t^*} (1 - F(t))dt - F(t^*) \int_0^{t^{**}} (1 - F(t))dt + E_{simple}.$$

Thus, we obtain

$$E_{simple} - E_{sched} = F(t^*) \int_0^{t^{**}} (1 - F(t))dt - F(t^{**}) \int_0^{t^*} (1 - F(t))dt =$$

$$F(t^*)F(t^{**}) \left( \frac{1}{F(t^{**})} \int_0^{t^{**}} (1 - F(t))dt - \frac{1}{F(t^*)} \int_0^{t^*} (1 - F(t))dt \right),$$

and since $t^*$ provides minimum for (53), the last expression is positive, which means that scheduling improves a simple restart policy.

Note, that we do not claim that the proposed scheduling policy is optimal – our example just shows that the pure restart strategy is not optimal. There should be an optimal combination interleaving restarts on the global level and scheduling on the local level, but finding this combination is left for future research.

## 9. Conclusions

In this work we present an algorithm for optimal scheduling of anytime algorithms with shared resources. We first introduce a formal framework for representing and analyzing scheduling strategies. We begin by analyzing the case where the only allowed scheduling operations are suspending and resuming processes. We prove necessary conditions for schedule optimality and present an algorithm for building optimal schedules that is based on those conditions. We then analyze the more general case where the scheduler can increase or decrease the intensity of the scheduled processes. We prove necessary conditions and show that intensity control is only rarely needed. We then analyze, theoretically and empirically, the behavior of our scheduling algorithm for various distribution types. Finally, we present empirical results of applying our scheduling algorithm to the Latin Square problem.

The results show that the optimal strategy indeed outperforms other scheduling strategies. For lognormal distribution, we showed an improvement of more than 50% over the naive sequential strategy. In general, our algorithm is particularly beneficial for heavy-tailed distributions, but even for exponential distribution we show a benefit of more than 35%.

In some cases, however, simple scheduling strategies yield results similar to those obtained by our algorithm. For example, the optimal schedule for uniform distribution is to apply one of the processes with no switch. When the probability to succeed within the given time limit approaches 1, this simple scheduling strategy also becomes close to optimal, at least for unimodal distributions with no strong skew towards zero. On the other hand, when the probability of success approaches zero, another simple strategy that applies the processes simultaneously becomes close to optimal.

Such a behavior meets the intuition. For heavy-tailed distributions, switching between the runs is promising because the chance to be on a bad trajectory is high enough. The same is correct for distributions with low probability of success. However, if the probability to be on a bad trajectory is too high, the best strategy is to switch between the runs as fast as possible, which is equivalent to the simultaneous strategy. On the other hand, if the distribution is too skewed to the right, often there is no sense to switch between the runs, since the new run should pay a high penalty before it reaches the "promising" distribution area. In general, when the user is certain that the particular application falls under one of the categories above, the cost of calculating the optimal schedule can be saved.

The high complexity of computation is one of the potential weaknesses of the presented algorithm. This complexity can be represented as a multiplication of three factors: function minimization, Branch-and-Bound search, and solving Equations (18) and (19) for the case of two agents or Equation (28) for the general case. For two agents, the only exponential component is the Branch-and-Bound search. We found, however, that in practice the branching factor, which is roughly the number of roots of the equations above, is rather small, while the depth of the search tree can be controlled by iterative-deepening strategies. For an arbitrary number of agents, function minimization may also be exponential. In practice, however, it depends on the behavior of the minimized function and the minimization algorithm.

Since the optimal schedule is static and can be applied to a large number of problem instances, its computation is beneficial even when associated with high cost. Moreover, in some applications (such as robotic search) the computational cost can be outweighed by the gain obtained from a single invocation.

The previous work most related to our research is the restart framework (Luby et al., 1993). The most important difference between our algorithm and the restart policy is the ability to handle the cases where the number of runs is limited, or where different algorithms are involved. When only one algorithm is available and the number of runs is infinite, the restart strategy is optimal. However, as we have shown in Section 8, some problems may benefit from the combination of these two approaches.

Our algorithm assumes the availability of the performance profiles of the processes. Such performance profiles can be derived analytically using theoretical models of the processes or empirically from previous experience with solving similar problems. Online learning of performance profiles, which could expand the applicability of the proposed framework, is a subject of ongoing research.

The framework presented here can be used for a wide range of applications. In the introduction we presented three examples. The first example describes two alternative learning algorithms working in parallel. The behavior of such algorithms is usually exponential, and the analysis for such setup is given in Section 6.2. The second example is a CSP problem with two alternative initial configurations, which is analogous to the Latin Square example of Sections 2.2 and 7. The last example includes crawling processes with a limited shared bandwidth. Unlike the first two examples, this setup falls under the framework of intensity control described in Section 5.

Similar schemes may be applied for more elaborate setups:

- Scheduling a system of $n$ anytime algorithms, where the overall cost of the system is defined as the maximal cost of its components (unlike the analysis in Section 4, this function is not differentiable);

- Scheduling with non-zero process switch costs;

- Providing dynamic scheduling algorithms able to handle changes in the environment;

- Building effective algorithms for the case of several resources of different types, e.g., multiprocessor systems.

## Appendix A. Formal Proofs

### A.1 Proof of Lemma 1

The claim of the lemma is as follows:

*For a system of $n$ processes, the expression for the expected cost (6) can be rewritten as*

$$E_u(\zeta_1, \ldots, \zeta_n, \ldots) = \sum_{k=0}^{\infty} \sum_{i=1}^{n} \prod_{j=i+1}^{i+n-1} (1 - F_{\#j}(\zeta_{k-1}^j)) \int_{\zeta_{k-1}^i}^{\zeta_k^i} (1 - F_i(x)) dx. \tag{55}$$

**Proof:** Splitting the whole integration range $[0, \infty)$ to the intervals $[t_k^{i-1}, t_k^i]$ yields the following expression:

$$E_u(\sigma_1, \ldots, \sigma_n) = \int_0^{\infty} \prod_{j=1}^{n} (1 - F_j(\sigma_j)) dt = \sum_{k=0}^{\infty} \sum_{i=1}^{n} \int_{t_k^{i-1}}^{t_k^i} \prod_{j=1}^{n} (1 - F_j(\sigma_j)) dt. \tag{56}$$

By (25), we can rewrite the inner integral as

$$\int_{t_k^{i-1}}^{t_k^i} \prod_{j=1}^{n} (1 - F_j(\sigma_j)) =$$

$$\int_{t_k^{i-1}}^{t_k^i} \left[ \prod_{j=1}^{i-1} (1 - F_j(\zeta_k^j)) \cdot (1 - F_i(t - t_k^{i-1} + \zeta_{k-1}^i)) \cdot \prod_{j=i+1}^{n} (1 - F_j(\zeta_{k-1}^j)) \right] dt = \tag{57}$$

$$\prod_{j=i+1-n}^{i-1} (1 - F_{\#j}(\zeta_k^j)) \int_{t_k^{i-1}}^{t_k^i} (1 - F_i(t - t_k^{i-1} + \zeta_{k-1}^i)) dt.$$

Substituting $x$ for $t - t_k^{i-1} + \zeta_{k-1}^i$ and using (23), we obtain

$$\prod_{j=i+1-n}^{i-1} (1 - F_{\#j}(\zeta_k^j)) \int_{t_k^{i-1}}^{t_k^i} (1 - F_i(t - t_k^{i-1} + \zeta_{k-1}^i)) dt =$$

$$\prod_{j=i+1}^{i+n-1} (1 - F_{\#j}(\zeta_{k-1}^j)) \int_{\zeta_{k-1}^i}^{t_k^i - t_k^{i-1} + \zeta_{k-1}^i} (1 - F_i(x)) dx = \tag{58}$$

$$\prod_{j=i+1}^{i+n-1} (1 - F_{\#j}(\zeta_{k-1}^j)) \int_{\zeta_{k-1}^i}^{\zeta_k^i} (1 - F_i(x)) dx.$$

Combining (56), (57) and (58) gives us (55).
*Q.E.D.*

## A.2 Proof of the Chain Theorem for $n$ Processes

The chain theorem claim is as follows:

*The value for $\zeta_{m+1}^{l-1}$ may either be $\zeta_m^{l-1}$, or can be computed given the previous $2n-2$ values of $\zeta$ using the formula*

$$\frac{f_l(\zeta_m^l)}{1 - F_l(\zeta_m^l)} = \frac{\displaystyle\prod_{j=l+1}^{l+n-1}(1 - F_{\#j}(\zeta_{m-1}^j)) - \prod_{j=l+1}^{l+n-1}(1 - F_{\#j}(\zeta_m^j))}{\displaystyle\sum_{i=l-n+1}^{l-1}\prod_{\substack{j=i+1 \\ \#j \neq l}}^{i+n-1}(1 - F_{\#j}(\zeta_m^j))\int_{\zeta_m^i}^{\zeta_{m+1}^i}(1 - F_{\#i}(x))dx} \tag{59}$$

**Proof:** By Lemma 1, the expression we want to minimize is described by the equation

$$E_u(\zeta_1, \ldots, \zeta_n, \ldots) = \sum_{k=0}^{\infty}\sum_{i=1}^{n}\prod_{j=i+1}^{i+n-1}(1 - F_{\#j}(\zeta_{k-1}^j))\int_{\zeta_{k-1}^i}^{\zeta_k^i}(1 - F_i(x))dx. \tag{60}$$

The expression above reaches its optimal values either when

$$\frac{dE_u}{d\zeta_j} = 0 \text{ for } j = 1, \ldots, n, \ldots, \tag{61}$$

or on the border described by (26).

Reaching the optimal values on the border corresponds to the first alternative described in the theorem. Let us now consider a case when the derivative of $E_u$ by $\zeta_j$ is 0.

Each variable $\zeta_j$ may be presented as $\zeta_{mn+l} = \zeta_m^l$, where $0 \leq l \leq n-1$. Let us see which summation terms of (60) $\zeta_m^l$ is participating in.

1. $\zeta_m^l$ may be a lower bound of the integral from (60). This happens when $k = m+1$ and $i = l$. The corresponding term is

$$S_0 = \prod_{j=l+1}^{l+n-1}(1 - F_{\#j}(\zeta_m^j))\int_{\zeta_m^l}^{\zeta_{m+1}^l}(1 - F_l(x))dx,$$

   and

$$\frac{dS_0}{d\zeta_m^l} = -(1 - F_l(\zeta_m^l)) \cdot \prod_{j=l+1}^{l+n-1}(1 - F_{\#j}(\zeta_m^j)).$$

2. $\zeta_m^l$ may be an upper bound of the same integral, which happens when $k = m$ and $i = l$. The corresponding term is

$$S_l = \prod_{j=l+1}^{l+n-1}(1 - F_{\#j}(\zeta_{m-1}^j))\int_{\zeta_{m-1}^l}^{\zeta_m^l}(1 - F_l(x))dx,$$

and

$$\frac{dS_l}{d\zeta_m^l} = (1 - F_l(\zeta_m^l)) \cdot \prod_{j=l+1}^{l+n-1} (1 - F_{\#j}(\zeta_{m-1}^j)).$$

3. Finally, $\zeta_m^l$ may participate in the product

$$\prod_{j=i+1}^{i+n-1} (1 - F_{\#j}(\zeta_{k-1}^j)).$$

For $i = 1 \ldots l-1$, this may happen when $k = m+1$ and $j = l$, and the corresponding term is

$$S_i = \prod_{j=i+1}^{i+n-1} (1 - F_{\#j}(\zeta_m^j)) \int_{\zeta_m^i}^{\zeta_{m+1}^i} (1 - F_i(x)) dx,$$

with the derivative

$$\frac{dS_i}{d\zeta_m^l} = -f_l(\zeta_m^l) \prod_{j=i+1, \#j \neq l}^{i+n-1} (1 - F_{\#j}(\zeta_m^j)) \int_{\zeta_m^i}^{\zeta_{m+1}^i} (1 - F_i(x)) dx.$$

For $i = l+1 \ldots n$, $k = m$ and $j = l+n$. The corresponding term is

$$S_i = \prod_{j=i+1}^{i+n-1} (1 - F_{\#j}(\zeta_{m-1}^j)) \int_{\zeta_{m-1}^i}^{\zeta_m^i} (1 - F_i(x)) dx,$$

with the derivative

$$\frac{dS_i}{d\zeta_m^l} = -f_l(\zeta_m^l) \prod_{j=i+1, \#j \neq l}^{i+n-1} (1 - F_{\#j}(\zeta_{m-1}^j)) \int_{\zeta_{m-1}^i}^{\zeta_m^i} (1 - F_i(x)) dx.$$

Since for $i = l$, $\zeta_m^l$ appears only in the integral, there is no other possibility for $\zeta_m^l$ to appear in the expression, and therefore

$$\frac{dE_u}{d\zeta_m^l} = \sum_{i=0}^{n} \frac{dS_i}{d\zeta_m^l}.$$

The right-hand side of the sum above can be written as follows:

$$\sum_{i=0}^{n} \frac{dS_i}{d\zeta_m^l} =$$

$$-(1 - F_l(\zeta_m^l)) \prod_{j=l+1}^{l+n-1} (1 - F_{\#j}(\zeta_m^j)) + (1 - F_l(\zeta_m^l)) \prod_{j=l+1}^{l+n-1} (1 - F_{\#j}(\zeta_{m-1}^j)) -$$

$$\sum_{i=1}^{l-1} f_l(\zeta_m^l) \prod_{j=i+1, \#j \neq l}^{i+n-1} (1 - F_{\#j}(\zeta_m^j)) \int_{\zeta_m^i}^{\zeta_{m+1}^i} (1 - F_i(x)) dx -$$

$$\sum_{i=l+1}^{n} f_l(\zeta_m^l) \prod_{j=i+1, \#j \neq l}^{i+n-1} (1 - F_{\#j}(\zeta_{m-1}^j)) \int_{\zeta_{m-1}^i}^{\zeta_m^i} (1 - F_i(x)) dx. \tag{62}$$

However,

$$\sum_{i=l+1}^{n} \prod_{j=i+1, \#j \neq l}^{i+n-1} (1 - F_{\#j}(\zeta_{m-1}^j)) \int_{\zeta_{m-1}^i}^{\zeta_m^i} (1 - F_i(x)) dx =$$
$$\sum_{i=l-n+1}^{0} \prod_{j=i+1, \#j \neq l}^{i+n-1} (1 - F_{\#j}(\zeta_m^j)) \int_{\zeta_m^i}^{\zeta_{m+1}^i} (1 - F_i(x)) dx. \tag{63}$$

Substituting (63) into (62), we obtain

$$\sum_{i=0}^{n} \frac{dS_i}{d\zeta_m^l} =$$
$$(1 - F_l(\zeta_m^l)) \left( \prod_{j=l+1}^{l+n-1} (1 - F_{\#j}(\zeta_{m-1}^j)) - \prod_{j=l+1}^{l+n-1} (1 - F_{\#j}(\zeta_m^j)) \right) -$$
$$f_l(\zeta_m^l) \sum_{i=l-n+1}^{l-1} \prod_{j=i+1, \#j \neq l}^{i+n-1} (1 - F_{\#j}(\zeta_m^j)) \int_{\zeta_m^i}^{\zeta_{m+1}^i} (1 - F_i(x)) dx. \tag{64}$$

If $1 - F_l(\zeta_m^l)$ were 0, that would mean that the goal has been reached with the probability of 1, and further scheduling would be redundant. Otherwise, expression in (64) is 0 when

$$\frac{f_l(\zeta_m^l)}{1 - F_l(\zeta_m^l)} = \frac{\displaystyle\prod_{j=l+1}^{l+n-1} (1 - F_{\#j}(\zeta_{m-1}^j)) - \prod_{j=l+1}^{l+n-1} (1 - F_{\#j}(\zeta_m^j))}{\displaystyle\sum_{i=l-n+1}^{l-1} \prod_{j=i+1, \#j \neq l}^{i+n-1} (1 - F_{\#j}(\zeta_m^j)) \int_{\zeta_m^i}^{\zeta_{m+1}^i} (1 - F_{\#i}(x)) dx},$$

which is equivalent to (59).

Equation (59) includes $2n - 1$ variables ($\zeta_{m-1}^{l+1} = \zeta_{n(m-1)+l+1}$ to $\zeta_{m+1}^{l-1} = \zeta_{n(m+1)+l-1}$), providing an implicit dependency of $\zeta_{m+1}^{l-1}$ on the remaining $2n - 2$ variables. $Q.E.D.$

## A.3 Proof of Lemma 2

The claim of the lemma is as follows:

*The Euler-Lagrange conditions for the minimization problem (33) yield two strong invariants:*

1. *For processes $k_1$ and $k_2$ for which $\sigma_{k_1}$ and $\sigma_{k_2}$ are not on the border described by (34), the distribution and density functions satisfy*

$$\frac{f_{k_1}(\sigma_{k_1})}{1 - F_{k_1}(\sigma_{k_1})} = \frac{f_{k_2}(\sigma_{k_2})}{1 - F_{k_2}(\sigma_{k_2})}. \tag{65}$$

2. *If the schedules of* all *the processes are not on the border described by (34), then either* $c = 1$ *or* $f_k(\sigma_k) = 0$ *for each k.*

**Proof:** Let $g(t, \sigma_1, \ldots, \sigma_n, \sigma'_1, \ldots, \sigma'_n)$ be the function under the integral sign of (33):

$$g(t, \sigma_1, \ldots, \sigma_n, \sigma'_1, \ldots, \sigma'_n) = \left( (1-c) + c \sum_{i=1}^{n} \sigma'_i \right) \prod_{j=1}^{n} (1 - F_j(\sigma_j)). \tag{66}$$

A necessary condition of Euler-Lagrange claims that a set of functions $\sigma_1, \ldots, \sigma_n$ provides a weak (local) minimum to the functional

$$E_u(\sigma_1, \ldots, \sigma_n) = \int_0^{\infty} g(t, \sigma_1, \ldots, \sigma_n, \sigma'_1, \ldots, \sigma'_n) dt$$

only if these functions satisfy a system of equations of the form

$$g'_{\sigma_k} - \frac{d}{dt} g'_{\sigma'_k} = 0. \tag{67}$$

In our case,

$$g'_{\sigma_k} = - \left( (1-c) + c \sum_{i=1}^{n} \sigma'_i \right) f_k(\sigma_k) \prod_{j \neq k} (1 - F_j(\sigma_j)), \tag{68}$$

and

$$\frac{d}{dt} g'_{\sigma'_k} = c \frac{d}{dt} \prod_{j=1}^{n} (1 - F_j(\sigma_j)) = -c \sum_{l=1}^{n} \sigma'_l f_l(\sigma_l) \prod_{j \neq l} (1 - F_j(\sigma_j)). \tag{69}$$

Substituting the last expression into (67), we obtain

$$g'_{\sigma_1} = g'_{\sigma_2} = \ldots = g'_{\sigma_n} = -c \sum_{l=1}^{n} \sigma'_l f_l(\sigma_l) \prod_{j \neq l} (1 - F_j(\sigma_j)),$$

and by (68) for every $k_1$ and $k_2$

$$f_{k_1}(\sigma_{k_1}) \prod_{j \neq k_1} (1 - F_j(\sigma_j)) = f_{k_2}(\sigma_{k_2}) \prod_{j \neq k_2} (1 - F_j(\sigma_j)).$$

We can ignore the case where one of the terms $1 - F_j(\sigma_j)$ is 0. Indeed, this is possible only if the goal is reached by process $j$ with probability of 1, and in this case no optimization is needed. Therefore, we obtain

$$f_{k_1}(\sigma_{k_1})(1 - F_{k_2}(\sigma_{k_2})) = f_{k_2}(\sigma_{k_2})(1 - F_{k_1}(\sigma_{k_1})), \tag{70}$$

which is equivalent to (65).

Let us show now the correctness of the second invariant. By (69) and (65), we obtain

$$
\frac{d}{dt} g'_{\sigma'_k} = -c \sum_{l=1}^{n} \sigma'_l f_l(\sigma_l) \prod_{j \neq l} (1 - F_j(\sigma_j)) =
$$

$$
- c \sum_{l=1}^{n} \sigma'_l \frac{f_l(\sigma_l)}{1 - F_l(\sigma_l)} \prod_{j=1}^{n} (1 - F_j(\sigma_j)) =
$$

$$
- c \sum_{l=1}^{n} \sigma'_l \frac{f_k(\sigma_k)}{1 - F_k(\sigma_k)} \prod_{j=1}^{n} (1 - F_j(\sigma_j)) =
$$

$$
- c \left( \sum_{i=1}^{n} \sigma'_i \right) f_k(\sigma_k) \prod_{j \neq k} (1 - F_j(\sigma_j)).
$$

By (36) we get

$$
g'_{\sigma_k} - \frac{d}{dt} g'_{\sigma'_k} = - \left( (1-c) + c \sum_{i=1}^{n} \sigma'_i \right) f_k(\sigma_k) \prod_{j \neq k} (1 - F_j(\sigma_j))
$$

$$
+ c \left( \sum_{i=1}^{n} \sigma'_i \right) f_k(\sigma_k) \prod_{j \neq k} (1 - F_j(\sigma_j)) =
$$

$$
- (1-c) f_k(\sigma_k) \prod_{j \neq k} (1 - F_j(\sigma_j)) = 0.
$$

Since we ignore the case when $(1 - F_j(\sigma_j)) = 0$, the second invariant is correct.
*Q.E.D.*

## A.4 Proof of Lemma 3

The claim of the lemma is as follows:

*If an optimal solution exists, then there exists an optimal solution $\sigma_1, \ldots, \sigma_n$, such that at each time $t$ all the resources are consumed, i.e.,*

$$
\forall t \sum_{i=1}^{n} \sigma'_i(t) = 1.
$$

*In the case where time cost is not zero $(c \neq 1)$, the equality above is a necessary condition for solution optimality.*

**Proof:** We know that $\{\sigma_i\}$ provide a minimum for the expression (33)

$$
\int_0^{\infty} \left( (1-c) + c \sum_{i=1}^{n} \sigma'_i \right) \prod_{j=1}^{n} (1 - F_j(\sigma_j)) dt.
$$

Let us assume that in some time interval $[t_0, t_1]$, $\{\sigma_i\}$ do not satisfy the lemma's constraints. However, it is possible to use the same amount of resources more effectively. Let us consider a linear time warp $\nu(t) = \alpha t + \beta$ on the time interval $[t_0, t_1]$, satisfying $\nu(t_0) = t_0$. From the last condition, it follows that $\beta = t_0(1 - \alpha)$. Let $t_1'$ be a point where $\nu(t)$ achieves $t_1$, i.e., $t_1' = t_0 + (t_1 - t_0)/\alpha$. Let us consider a set of new objective schedule functions $\widetilde{\sigma}_i(t)$ of the form

$$\widetilde{\sigma}_i(t) = \begin{cases} \sigma_i(t), & t \leq t_0, \\ \sigma_i(\alpha t + \beta), & t_0 \leq t \leq t_1', \\ \sigma_i(t + t_1 - t_1'), & t > t_1'. \end{cases}$$

Thus, $\widetilde{\sigma}_i(t)$ behaves as $\sigma_i(t)$ before $t_0$, as $\sigma_i(t)$ with a time shift after $t_1'$, and as a linearly speeded up version of $\sigma_i(t)$ in the interval $[t_0, t_1']$. Since $\nu(t_0) = t_0$ and $\nu(t_1') = t_1$, $\widetilde{\sigma}_i(t)$ is continuous at the points $t_0$ and $t_1'$.

$\widetilde{\sigma}_i'(t)$ is equal to $\alpha \sigma_i'(t)$ within the interval $[t_0, t_1]$, and to $\sigma_i'(t)$ outside this interval. By the contradiction assumption, $\sigma_i$ do not meet the lemma constraints in $[t_0, t_1]$, and thus we can take

$$\alpha = \frac{1}{\max_{t \in [t_0, t_1]} \sum_{i=1}^{n} \sigma_i'(t)} > 1,$$

leading to valid functions $\widetilde{\sigma}_i'(t)$. Using $\widetilde{\sigma}_i(t)$ in (33), we obtain

$$\begin{aligned} E_u(\widetilde{\sigma}_1, \ldots, \widetilde{\sigma}_n) &= \int_0^\infty \left( (1-c) + c\sum_{i=1}^{n} \widetilde{\sigma}_i'(t) \right) \prod_{j=1}^{n} (1 - F_j(\widetilde{\sigma}_j(t))) dt = \\ &\int_0^{t_0} \left( (1-c) + c\sum_{i=1}^{n} \sigma_i'(t) \right) \prod_{j=1}^{n} (1 - F_j(\sigma_j(t))) dt + \\ &\int_{t_0}^{t_1'} \left( (1-c) + c\alpha \sum_{i=1}^{n} \sigma_i'(\alpha t + \beta) \right) \prod_{j=1}^{n} (1 - F_j(\sigma_j(\alpha t + \beta))) dt + \\ &\int_{t_1'}^{\infty} \left( (1-c) + c\sum_{i=1}^{n} \sigma_i'(t + t_1 - t_1') \right) \prod_{j=1}^{n} (1 - F_j(\sigma_j(t + t_1 - t_1'))) dt. \end{aligned}$$

By substituting $x = \alpha t + \beta$ in the second term of the last sum, and $x = t + t_1 - t_1'$ in the third term, we obtain

$$\begin{aligned} E_u(\widetilde{\sigma}_1, \ldots, \widetilde{\sigma}_n) &= \int_0^{t_0} \left( (1-c) + c\sum_{i=1}^{n} \sigma_i'(t) \right) \prod_{j=1}^{n} (1 - F_j(\sigma_j(t))) dt + \\ &\int_{t_0}^{t_1} \left( \frac{1-c}{\alpha} + c\sum_{i=1}^{n} \sigma_i'(x) \right) \prod_{j=1}^{n} (1 - F_j(\sigma_j(x))) dx + \\ &\int_{t_1}^{\infty} \left( (1-c) + c\sum_{i=1}^{n} \sigma_i'(x) \right) \prod_{j=1}^{n} (1 - F_j(\sigma_j(x))) dx = \\ &E_u(\sigma_1, \ldots, \sigma_n) - \int_{t_0}^{t_1} (1-c) \left( 1 - \frac{1}{\alpha} \right) \prod_{j=1}^{n} (1 - F_j(\sigma_j)) dt. \end{aligned}$$

Since $\alpha > 1$, the last term is non-negative, and therefore

$$E_u(\widetilde{\sigma}_1, \ldots, \widetilde{\sigma}_n) \leq E_u(\sigma_1, \ldots, \sigma_n),$$

meaning that the set $\{\tilde{\sigma}_i\}$ provides a solution of at least the same quality as $\{\sigma_i\}$. If $c \neq 1$, this contradicts to the optimality of the original schedule, and if $c = 1$, the new schedule will also be optimal.

*Q.E.D.*

## A.5 Proof of Theorem 4

The claim of the theorem is as follows:

*Let the set of functions $\{\sigma_i\}$ be a solution of minimization problem (6) under constraints (34). Let $t_0$ be a point where the hazard functions of all the processes $h_i(\sigma_i(t))$ are continuous, and let $A_k$ be the process active at $t_0$ ($\sigma'_k(t_0) > 0$), such that for any other process $A_i$*

$$h_i(\sigma_i(t_0)) < h_k(\sigma_k(t_0)). \tag{71}$$

*Then at $t_0$ process $k$ consumes all the resources, i.e. $\sigma'_k(t_0) = 1$.*

**Proof:** First we want to prove the theorem for the case of two processes, and then to generalize the proof to the case of $n$ processes. Assume that $\sigma_1(t)$ and $\sigma_2(t)$ provide the optimal solution, and at some point $t_0$ $\sigma'_1(t_0) > 0$ and

$$\frac{f_1(\sigma_1(t_0))}{1 - F_1(\sigma_1(t_0))} > \frac{f_2(\sigma_2(t_0))}{1 - F_2(\sigma_2(t_0))}. \tag{72}$$

From the continuity of the functions $h_i(t)$ in the point $t_0$, it follows that there exists some neighborhood $U(t_0)$ of $t_0$, such that for each two points $t', t''$ in this neighborhood $h_1(t') > h_2(t'')$, i.e.,

$$\min_{t' \in U(t_0)} \frac{f_1(\sigma_1(t'))}{1 - F_1(\sigma_1(t'))} > \max_{t'' \in U(t_0)} \frac{f_2(\sigma_2(t''))}{1 - F_2(\sigma_2(t''))}. \tag{73}$$

Let us consider some interval $[t_0, t_1] \subset U(t_0)$. In order to make the proof more readable, we introduce the following notation (for this proof only):

- We denote $\sigma_1(t)$ by $\sigma(t)$. By Lemma 3, $\sigma_2(t) = t - \sigma(t)$.

- We denote $\sigma(t_0)$ by $\sigma^0$ and $\sigma(t_1)$ by $\sigma^1$.

In the interval $[t_0, t_1]$ the first process obtains $\sigma^1 - \sigma^0$ resources, and the second process obtains $(t_1 - t_0) - (\sigma^1 - \sigma^0)$ resources. Let us consider a special resource distribution $\tilde{\sigma}$, which first gives all the resources to the first process, and then to the second process, keeping the same quantity of resources as $\sigma$:

$$\tilde{\sigma}(t) = \begin{cases} \sigma(t), & t \leq t_0, \\ t - t_0 + \sigma^0, & t_0 \leq t \leq t_0 + \sigma^1 - \sigma^0, \\ \sigma^1, & t_0 + \sigma^1 - \sigma^0 \leq t \leq t_1 \\ \sigma(t), & t \geq t_1. \end{cases}$$

It is easy to see that $\sigma(t)$ is continuous at the points $t_0$, $t_1$, and $t_0 + \sigma^1 - \sigma^0$. We want to show that, unless the first process consumes all the resources at the beginning, the schedule produced by $\tilde{\sigma}$ outperforms the schedule produced by $\sigma$.

Let $t^* = t_0 + \sigma^1 - \sigma^0$, which corresponds to the time when the first process would have consumed all its resources had it been working with the maximal intensity. First, we want to show that in the interval $[t_0, t^*]$

$$(1 - F_1(\sigma(t)))(1 - F_2(t - \sigma(t))) \geq (1 - F_1(t - t_0 + \sigma^0))(1 - F_2(t_0 - \sigma^0)). \tag{74}$$

Let

$$\nu(t) = (t - t_0 + \sigma^0) - \sigma(t). \tag{75}$$

The inequality (74) becomes

$$(1 - F_1(t - t_0 + \sigma^0 - \nu(t)))(1 - F_2(t_0 - \sigma^0 + \nu(t))) \geq (1 - F_1(t - t_0 + \sigma^0))(1 - F_2(t_0 - \sigma^0)). \tag{76}$$

Let us find a value of $x = \nu(t)$ that provides the minimum to the left-hand side of (76) for the fixed $t$. Let us denote

$$G(x) = (1 - F_1(t - t_0 + \sigma^0 - x))(1 - F_2(t_0 - \sigma^0 + x)).$$

Then,

$$G'(x) = f_1(t - t_0 + \sigma^0 - x)(1 - F_2(t_0 - \sigma^0 + x)) - f_2(t_0 - \sigma^0 + x)(1 - F_1(t - t_0 + \sigma^0 - x)).$$

Since a valid $\sigma(t)$ in the interval $[t_0, t_1]$ obtains values between $\sigma^0$ and $\sigma^1$, by (75) we have

$$t - t_0 + \sigma^0 - x \in [\sigma^0, \sigma^1],$$
$$t_0 - \sigma^0 + x \in [t_0 - \sigma^0, t_1 - \sigma^1].$$

Therefore, there exist $t', t'' \in [t_0, t_1]$, such that $\sigma_1(t') = \sigma(t') = t - t_0 + \sigma^0 - x$ and $\sigma_2(t'') = t'' - \sigma(t'') = t_0 - \sigma^0 + x$. By (73) we obtain $G'(x) > 0$, meaning that $G(x)$ monotonically increases. Besides, by (75) we have $x = \nu(t) \geq 0$ (since $\sigma'(t) \leq 1$), and therefore $G(x)$ obtains its minimal value when $x = 0$. Therefore, if we denote by $Ran(t)$ the set of valid values for $\nu(t)$,

$$(1 - F_1(\sigma))(1 - F_2(t - \sigma)) = (1 - F_1(t - t_0 + \sigma^0 - \nu(t)))(1 - F_2(t_0 - \sigma^0 + \nu(t))) \geq$$
$$\min_{x \in Ran(t)} (1 - F_1(t - t_0 + \sigma^0 - x))(1 - F_2(t_0 - \sigma^0 + x)) =$$
$$(1 - F_1(t - t_0 + \sigma^0))(1 - F_2(t_0 - \sigma^0)),$$

and the strict equality occurs if and only if $\sigma(t) = t - t_0 + \sigma^0$. Thus,

$$(1 - F_1(\sigma))(1 - F_2(t - \sigma)) \geq (1 - F_1(\tilde{\sigma}))(1 - F_2(t - \tilde{\sigma}))$$

for $t \in [t_0, t^*]$.

Let us show now the correctness of the same statement in the interval $[t^*, t_1]$, which is equivalent to the inequality

$$(1 - F_1(\sigma(t)))(1 - F_2(t - \sigma(t))) \geq (1 - F_1(\sigma^1))(1 - F_2(t - \sigma^1)). \tag{77}$$

The proof is similar. Let

$$\nu(t) = \sigma^1 - \sigma(t). \tag{78}$$

The inequality (77) becomes

$$(1 - F_1(\sigma^1 - \nu(t)))(1 - F_2(t - \sigma^1 + \nu(t))) \geq (1 - F_1(\sigma^1))(1 - F_2(t - \sigma^1)). \tag{79}$$

As before, we find a value of $x = \nu(t)$ that provides the minimum to the left-hand side of (79)

$$G(x) = (1 - F_1(\sigma^1 - x))(1 - F_2(t - \sigma^1 + x)).$$

The derivative of $G(x)$ is

$$G'(x) = f_1(\sigma^1 - x)(1 - F_2(t - \sigma^1 + x)) - f_2(t - \sigma^1 + x)(1 - F_1(\sigma^1 - x)),$$

and since a valid $\sigma(t)$ in the interval $[t_0, t_1]$ obtains values between $\sigma^0$ and $\sigma^1$, by (78) we have

$$\sigma^1 - x \in [\sigma^0, \sigma^1],$$
$$t - \sigma^1 + x \in [t_0 - \sigma^0, t_1 - \sigma^1].$$

Therefore, there exist $t', t'' \in [t_0, t_1]$, such that $\sigma_1(t') = \sigma(t') = \sigma^1 - x$ and $\sigma_2(t'') = t'' - \sigma(t'') = t - \sigma^1 + x$. By (73), $G'(x) > 0$, and therefore $G(x)$ monotonically increases. Since $x = \sigma^1 - \sigma(t) \geq 0$, $G(x) \geq G(0)$. Thus, for $t \in [t^*, t_1]$,

$$(1 - F_1(\sigma))(1 - F_2(t - \sigma)) = (1 - F_1(\sigma^1 - \nu(t)))(1 - F_2(t - \sigma^1 + \nu(t))) \geq$$
$$\min_{x \in Ran(t)} (1 - F_1(\sigma^1 - x))(1 - F_2(t - \sigma^1 + x)) = (1 - F_1(\sigma^1))(1 - F_2(t - \sigma^1)),$$

and the strict equality occurs if and only if $\sigma(t) = \sigma^1$.

Combining this result with the previous one, we obtain that

$$(1 - F_1(\sigma))(1 - F_2(t - \sigma)) \geq (1 - F_1(\widetilde{\sigma}))(1 - F_2(t - \widetilde{\sigma}))$$

holds for every $t \in [t_0, t_1]$. Since $\widetilde{\sigma}(t)$ behaves as $\sigma(t)$ outside this interval, $E_u(\sigma) \geq E_u(\widetilde{\sigma})$. Besides, since the equality is obtained if and only if $\sigma \equiv \widetilde{\sigma}$, and since $E_u(\sigma)$ is optimal, we obtain that $\sigma \equiv \widetilde{\sigma}$, and therefore the first process will take all the resources in some interval $[t_0, t_1]$.

The proof for $n$ processes is exactly the same. Let $\{\sigma_i\}$ provide the optimal solution, and at the point $t_0$ there is process $k$, such that for each $j \neq k$

$$h_k(\sigma_k(t_0)) > h_j(\sigma_j(t_0)).$$

From the continuity of the functions $h_i(\sigma_i(t))$ in the point $t_0$, it follows that there exists some neighborhood $U(t_0)$ of $t_0$, such that

$$\min_{t' \in U(t_0)} h_k(\sigma_k(t')) > \max_{i \neq k} \max_{t'' \in U(t_0)} h_i(\sigma_i(t'')). \tag{80}$$

Let us take any process $l \neq k$, and let

$$y(t) = \sigma_k(t) + \sigma_l(t).$$

Now we can repeat the above proof while substituting $y(t)$ instead of $t$ under the function sign:

$$\widetilde{\sigma}_k(t) = \begin{cases} \sigma_k(t), & y(t) \leq y(t_0), \\ y(t) - y(t_0) + \sigma_k(t_0), & y(t_0) \leq y(t) \leq y(t_0) + \sigma_k(t_1) - \sigma_k(t_0), \\ \sigma_k(t_1), & y(t_0) + \sigma_k(t_1) - \sigma_k(t_0) \leq y(t) \leq y(t_1), \\ \sigma_k(t), & y(t) \geq y(t_1). \end{cases}$$

The substitution above produces a valid schedule due to the monotonicity of $y(t)$. The rest of the proof remains unchanged.

*Q.E.D.*

### A.6 Proof of Theorem 5

The claim of the theorem is as follows:

*An active process will remain active and consume all resources as long as its hazard function is monotonically increasing.*

**Proof:** The proof is by contradiction. Let $\{\sigma_j\}$ form an optimal schedule. Assume that at some point $t_1$ process $A_k$ is suspended, while its hazard function $h_k(\sigma_k(t_1))$ is monotonically increasing at $t_1$.

Let us assume first that at some point $t_2$ process $A_k$ becomes active again. Since we do not consider the case of making process active at a single point, there exists some $\Delta > 0$, such that $A_k$ is active in the intervals $[t_1 - \Delta, t_1]$ and $[t_2, t_2 + \Delta]$. $A_k$ has been stopped at a point of monotonicity of the hazard function, and therefore, by Theorem 4, in these intervals $A_k$ is the only active process. We consider two alternative scenarios. In the first one, we allow $A_k$ to be active for additional $\Delta$ time starting at $t_1$ (i.e., shifting its idle period by $\Delta$), while in the second we suspend $A_k$ by $\Delta$ earlier.

For the first scenario, the scheduling functions have the following form:

$$\sigma_k^a(t) = \begin{cases} \sigma_k(t), & t \leq t_1, \\ \sigma_k(t_1) + (t - t_1), & t_1 \leq t \leq t_1 + \Delta, \\ \sigma_k(t - \Delta) + \Delta = \sigma_k(t_1) + \Delta, & t_1 + \Delta \leq t \leq t_2 + \Delta, \\ \sigma_k(t), & t \geq t_2 + \Delta; \end{cases} \tag{81}$$

$$\sigma_j^a(t) = \begin{cases} \sigma_j(t), & t \leq t_1, \\ \sigma_j(t_1), & t_1 \leq t \leq t_1 + \Delta, \\ \sigma_j(t - \Delta), & t_1 + \Delta \leq t \leq t_2 + \Delta, \\ \sigma_j(t), & t \geq t_2 + \Delta. \end{cases} \tag{82}$$

It is possible to see that these scheduling functions are continuous and satisfy invariant (39), which makes this set a suitable candidate for optimality.

Substituting these values of $\sigma$ into (6), we obtain

$$E_u(\sigma_1^a, \ldots, \sigma_n^a) = \int_0^{t_1} \prod_{j=1}^n (1 - F_j(\sigma_j(t)))dt +$$

$$\int_{t_1}^{t_1+\Delta} (1 - F_k(\sigma_k(t_1) + (t - t_1))) \prod_{j \neq k} (1 - F_j(\sigma_j(t_1)))dt +$$

$$\int_{t_1+\Delta}^{t_2+\Delta} (1 - F_k(\sigma_k(t_1) + \Delta)) \prod_{j \neq k} (1 - F_j(\sigma_j(t - \Delta)))dt + \int_{t_2+\Delta}^{\infty} \prod_{j=1}^n (1 - F_j(\sigma_j(t)))dt =$$

$$\int_0^{t_1} \prod_{j=1}^n (1 - F_j(\sigma_j(t)))dt + \prod_{j \neq k} (1 - F_j(\sigma_j(t_1))) \int_0^{\Delta} (1 - F_k(\sigma_k(t_1) + x))dx +$$

$$\int_{t_1}^{t_2} (1 - F_k(\sigma_k(t_1) + \Delta)) \prod_{j \neq k} (1 - F_j(\sigma_j(t)))dt + \int_{t_2+\Delta}^{\infty} \prod_{j=1}^n (1 - F_j(\sigma_j(t)))dt.$$

Subtracting $E_u(\sigma_1, \ldots, \sigma_n)$ given by (6) from $E_u(\sigma_1^a, \ldots, \sigma_n^a)$, we get

$$E_u(\sigma_1, \ldots, \sigma_n) - E_u(\sigma_1^a, \ldots, \sigma_n^a) =$$

$$\int_{t_1}^{t_2} [(1 - F_k(\sigma_k(t))) - (1 - F_k(\sigma_k(t_1) + \Delta))] \prod_{j \neq k} (1 - F_j(\sigma_j(t)))dt + \tag{83}$$

$$\int_{t_2}^{t_2+\Delta} \prod_{j=1}^n (1 - F_j(\sigma_j(t)))dt - \prod_{j \neq k} (1 - F_j(\sigma_j(t_1))) \int_0^{\Delta} (1 - F_k(\sigma_k(t_1) + x))dx.$$

Let us consider the first term of the last equation. Since in the interval $[t_1, t_2]$ $\sigma_k(t) = \sigma_k(t_1)$, in this interval

$$(1 - F_k(\sigma_k(t))) - (1 - F_k(\sigma_k(t_1) + \Delta)) = (1 - F_k(\sigma_k(t_1))) - (1 - F_k(\sigma_k(t_1) + \Delta)) =$$

$$-\int_0^{\Delta} d(1 - F_k(\sigma_k(t_1) + x)) = \int_0^{\Delta} f_k(\sigma_k(t_1) + x)dx = \int_0^{\Delta} h_k(\sigma_k(t_1) + x)(1 - F_k(\sigma_k(t_1) + x))dx.$$

Due to monotonicity of $h_k(\sigma_k)$ in $t_1$,

$$(1 - F_k(\sigma_k(t))) - (1 - F_k(\sigma_k(t_1) + \Delta)) =$$

$$\int_0^{\Delta} h_k(\sigma_k(t_1) + x)(1 - F_k(\sigma_k(t_1) + x))dx > h_k(\sigma_k(t_1)) \int_0^{\Delta} (1 - F_k(\sigma_k(t_1) + x))dx,$$

which leads to

$$\int_{t_1}^{t_2} [(1 - F_k(\sigma_k(t))) - (1 - F_k(\sigma_k(t_1) + \Delta))] \prod_{j \neq k} (1 - F_j(\sigma_j(t)))dt >$$

$$h_k(\sigma_k(t_1)) \int_0^{\Delta} (1 - F_k(\sigma_k(t_1) + x))dx \int_{t_1}^{t_2} \prod_{j \neq k} (1 - F_j(\sigma_j(t)))dt. \tag{84}$$

Let us now consider the second term of (83). Since in the interval $[t_2, t_2 + \Delta]$ only $A_k$ is active, in this interval

$$\sigma_j(t) = \begin{cases} \sigma_j(t_2), & j \neq k, \\ \sigma_k(t_1) + (t - t_2), & j = k. \end{cases}$$

Thus,

$$\int_{t_2}^{t_2+\Delta} \prod_{j=1}^n (1 - F_j(\sigma_j(t)))dt = \prod_{j \neq k}(1 - F_j(\sigma_j(t_2))) \int_0^\Delta (1 - F_k(\sigma_k(t_1) + x))dx. \quad (85)$$

Substituting (84) and (85) into (83), we obtain

$$E_u(\sigma_1, \ldots, \sigma_n) - E_u(\sigma_1^a, \ldots, \sigma_n^a) > \int_0^\Delta (1 - F_k(\sigma_k(t_1) + x))dx \times$$

$$\left[ h_k(\sigma_k(t_1)) \int_{t_1}^{t_2} \prod_{j \neq k}(1 - F_j(\sigma_j(t)))dt + \prod_{j \neq k}(1 - F_j(\sigma_j(t_2))) - \prod_{j \neq k}(1 - F_j(\sigma_j(t_1))) \right]. \quad (86)$$

The proof for the second scenario, where $A_k$ is suspended for $\Delta$, is similar. For this scenario, the scheduling functions $\sigma_k(t)$ and $\sigma_j(t)$ for $j \neq k$ can be represented as follows:

$$\sigma_k^i(t) = \begin{cases} \sigma_k(t), & t \leq t_1 - \Delta, \\ \sigma_k(t_1 - \Delta) = \sigma_k(t_1) - \Delta, & t_1 - \Delta \leq t \leq t_2 - \Delta, \\ \sigma_k(t_1 - \Delta) + (t - (t_2 - \Delta)) = \sigma_k(t_1) + (t - t_2), & t_2 - \Delta \leq t \leq t_2, \\ \sigma_k(t), & t \geq t_2; \end{cases} \quad (87)$$

$$\sigma_j^i(t) = \begin{cases} \sigma_j(t), & t \leq t_1 - \Delta, \\ \sigma_j(t + \Delta), & t_1 - \Delta \leq t \leq t_2 - \Delta, \\ \sigma_j(t_2), & t_2 - \Delta \leq t \leq t_2, \\ \sigma_j(t), & t \geq t_2. \end{cases} \quad (88)$$

As before, these scheduling functions are continuous and satisfy invariant (39).

Substituting $\sigma^i$ into (6), we obtain

$$E_u(\sigma_1^i, \ldots, \sigma_n^i) = \int_0^{t_1-\Delta} \prod_{j=1}^n (1 - F_j(\sigma_j(t)))dt +$$

$$\int_{t_1-\Delta}^{t_2-\Delta} (1 - F_k(\sigma_k(t_1) - \Delta)) \prod_{j \neq k}(1 - F_j(\sigma_j(t + \Delta)))dt +$$

$$\int_{t_2-\Delta}^{t_2} (1 - F_k(\sigma_k(t_1) + (t - t_2))) \prod_{j \neq k}(1 - F_j(\sigma_j(t_2)))dt + \int_{t_2}^\infty \prod_{j=1}^n (1 - F_j(\sigma_j(t)))dt =$$

$$\int_0^{t_1-\Delta} \prod_{j=1}^n (1 - F_j(\sigma_j(t)))dt + \prod_{j \neq k}(1 - F_j(\sigma_j(t_2))) \int_0^\Delta (1 - F_k(\sigma_k(t_1) - x))dx +$$

$$\int_{t_1}^{t_2} (1 - F_k(\sigma_k(t_1) - \Delta)) \prod_{j \neq k}(1 - F_j(\sigma_j(t)))dt + \int_{t_2+\Delta}^\infty \prod_{j=1}^n (1 - F_j(\sigma_j(t)))dt.$$

Subtracting $E_u(\sigma_1, \ldots, \sigma_n)$ given by (6) from $E_u(\sigma_1^i, \ldots, \sigma_n^i)$, we get

$$E_u(\sigma_1, \ldots, \sigma_n) - E_u(\sigma_1^i, \ldots, \sigma_n^i) =$$
$$\int_{t_1}^{t_2} [(1 - F_k(\sigma_k(t))) - (1 - F_k(\sigma_k(t_1) - \Delta))] \prod_{j \neq k}(1 - F_j(\sigma_j(t)))dt +$$
$$\int_{t_1 - \Delta}^{t_1} \prod_{j=1}^{n}(1 - F_j(\sigma_j(t)))dt - \prod_{j \neq k}(1 - F_j(\sigma_j(t_2))) \int_{0}^{\Delta}(1 - F_k(\sigma_k(t_1) - x))dx. \quad (89)$$

As in the first scenario, in the interval $[t_1, t_2]$

$$(1 - F_k(\sigma_k(t))) - (1 - F_k(\sigma_k(t_1) - \Delta)) = (1 - F_k(\sigma_k(t_1))) - (1 - F_k(\sigma_k(t_1) - \Delta)) =$$
$$\int_{-\Delta}^{0} d(1 - F_k(\sigma_k(t_1) + x)) = -\int_{-\Delta}^{0} f_k(\sigma_k(t_1) + x)dx =$$
$$-\int_{0}^{\Delta} f_k(\sigma_k(t_1) - x)dx = -\int_{0}^{\Delta} h_k(\sigma_k(t_1) - x)(1 - F_k(\sigma_k(t_1) - x))dx.$$

Due to monotonicity of $h_k(\sigma_k)$ in $t_1$,

$$(1 - F_k(\sigma_k(t))) - (1 - F_k(\sigma_k(t_1) - \Delta)) =$$
$$-\int_{0}^{\Delta} h_k(\sigma_k(t_1) - x)(1 - F_k(\sigma_k(t_1) - x))dx > -h_k(\sigma_k(t_1)) \int_{0}^{\Delta}(1 - F_k(\sigma_k(t_1) - x))dx,$$

which leads to

$$\int_{t_1}^{t_2} [(1 - F_k(\sigma_k(t))) - (1 - F_k(\sigma_k(t_1) - \Delta))] \prod_{j \neq k}(1 - F_j(\sigma_j(t)))dt >$$
$$-h_k(\sigma_k(t_1)) \int_{0}^{\Delta}(1 - F_k(\sigma_k(t_1) - x))dx \int_{t_1}^{t_2} \prod_{j \neq k}(1 - F_j(\sigma_j(t)))dt. \quad (90)$$

The transformations for the second term of (89) are also similar to the previous scenario. Since in the interval $[t_1 - \Delta, t_1]$ only $A_k$ is active, in this interval

$$\sigma_j(t) = \begin{cases} \sigma_j(t_1), & j \neq k, \\ \sigma_k(t_1) - (t_1 - t), & j = k. \end{cases}$$

Thus,

$$\int_{t_1 - \Delta}^{t_1} \prod_{j=1}^{n}(1 - F_j(\sigma_j(t)))dt = \prod_{j \neq k}(1 - F_j(\sigma_j(t_1))) \int_{0}^{\Delta}(1 - F_k(\sigma_k(t_1) - x))dx. \quad (91)$$

Substituting (90) and (91) into (89), we obtain

$$E_u(\sigma_1, \ldots, \sigma_n) - E_u(\sigma_1^i, \ldots, \sigma_n^i) > \int_{0}^{\Delta}(1 - F_k(\sigma_k(t_1) - x))dx \times$$
$$-\left[ h_k(\sigma_k(t_1)) \int_{t_1}^{t_2} \prod_{j \neq k}(1 - F_j(\sigma_j(t)))dt + \prod_{j \neq k}(1 - F_j(\sigma_j(t_2))) - \prod_{j \neq k}(1 - F_j(\sigma_j(t_1))) \right].$$
$$(92)$$

By (86) and (92),

$$\text{sign}(E_u(\sigma_1, \ldots, \sigma_n) - E_u(\sigma_1^a, \ldots, \sigma_n^a)) = -\text{sign}(E_u(\sigma_1, \ldots, \sigma_n) - E_u(\sigma_1^i, \ldots, \sigma_n^i)), \quad (93)$$

and therefore one of these scenarios leads to better schedule, which contradicts the optimality of the original one.

The proof for the case where control does not return to $A_k$ at all is exactly the same and is omitted here. Informally, it can be viewed as replacing $t_2$ by $\infty$ in all the formulas above, and the results are the same. same results.

*Q.E.D.*

## A.7 Proof of Theorem 6

The claim of the theorem is as follows:

*If no time cost is taken into account ($c = 1$), the model with shared resources under intensity control settings is equivalent to the model with independent processes under suspend-resume control settings. Namely, given a suspend-resume solution for the model with independent processes, we may reconstruct an intensity-based solution with the same cost for the model with shared resources and vice versa.*

**Proof:** Let $E_{shared}^*$ be the optimal value for the framework with shared resources, and $E_{independent}^*$ be the optimal value for the framework with independent processes. Since $c = 1$, the two problems minimize the same expression

$$E_u(\sigma_1, \ldots, \sigma_n) = \int_0^\infty \left( \sum_{i=1}^n \sigma_i' \right) \prod_{j=1}^n (1 - F_j(\sigma_j)) dt \to \min, \quad (94)$$

and each set $\{\sigma_i\}$ satisfying the resource sharing constraints automatically satisfies the process independence constraints, we obtain

$$E_{independent}^* \leq E_{shared}^*.$$

Let us prove that

$$E_{shared}^* \leq E_{independent}^*.$$

Assume that a set of functions $\sigma_1, \sigma_2, \ldots, \sigma_n$ is an optimal solution for the problem with independent processes, i.e.,

$$E_u(\sigma_1, \ldots, \sigma_n) = E_{independent}^*.$$

We want to construct a set of functions $\{\widetilde{\sigma_i}\}$ satisfying the resource sharing constrains, such that

$$E_u(\widetilde{\sigma_1}, \ldots, \widetilde{\sigma_n}) = E_u(\sigma_1, \ldots, \sigma_n).$$

Let us consider a set of discontinuity points of $\sigma_i'$

$$\mathcal{T} = \{t | \exists i : \sigma_i'(t - \epsilon) \neq \sigma_i'(t + \epsilon)\}.$$

In our model this set is countable, and we can write it as a sorted sequence $t_0 = 0 < t_1 < \ldots < t_k < \ldots$. The expected schedule cost in this case will have a form

$$E_u(\sigma_1, \ldots, \sigma_n) = \sum_{j=0}^{\infty} E_{u_j}(\sigma_1, \ldots, \sigma_n),$$

where

$$E_{u_j}(\sigma_1, \ldots, \sigma_n) = \int_{t_j}^{t_{j+1}} \left( \sum_{i=1}^{n} \sigma_i' \right) \prod_{l=1}^{n} (1 - F_l(\sigma_l)) dt.$$

We want to construct the functions $\widetilde{\sigma_i}$ incrementally. For each time interval $[t_j, t_{j+1}]$ we define a corresponding point $\widetilde{t_j}$ and a set of functions $\widetilde{\sigma_i}$, such that

$$\widetilde{E_{u_j}}(\widetilde{\sigma_1}, \ldots, \widetilde{\sigma_n}) = \int_{\widetilde{t_j}}^{\widetilde{t_{j+1}}} \left( \sum_{l=1}^{n} \widetilde{\sigma_l}' \right) \prod_{l=1}^{n} (1 - F_l(\widetilde{\sigma_l})) dt = E_{u_j}(\sigma_1, \ldots \sigma_n).$$

Let us denote $\sigma_{ij} = \sigma_i(t_j)$ and $\widetilde{\sigma_{ij}} = \widetilde{\sigma_i}(t_j)$. At the beginning, $\widetilde{\sigma_{i0}} = 0$ for each $i$, and $\widetilde{t_0} = 0$. Assume now that we have $\widetilde{t_{j'}}$ defined for $j' < j$, and $\widetilde{\sigma_i}(t)$ defined on each interval $[\widetilde{t_{j'}}, \widetilde{t_{j'+1}}]$. Let us show how to define $\widetilde{t_j}$ and $\widetilde{\sigma_j}$ on $[\widetilde{t_j}, \widetilde{t_{j+1}}]$.

By definition of $t_j$, $k = \sum_{l=1}^{n} \sigma_l'(t)$ is a constant for $t \in [t_j, t_{j+1}]$. Since $\{\sigma_i\}$ satisfy suspend-resume constraints, exactly $k \geq 1$ processes are active in this interval, each with full intensity. Without loss of generality, the active processes are $A_1, A_2, \ldots, A_k$, and

$$E_{u_j}(\sigma_1, \ldots, \sigma_n) = k \int_{t_j}^{t_{j+1}} \prod_{l=1}^{n} (1 - F_l(\sigma_l)) dt =$$

$$k \prod_{l=k+1}^{n} (1 - F_l(\sigma_{lj})) \int_{t_j}^{t_{j+1}} \prod_{l=1}^{k} (1 - F_l(t - t_j + \sigma_{lj})) dt =$$

$$k \prod_{l=k+1}^{n} (1 - F_l(\sigma_{lj})) \int_{0}^{t_{j+1} - t_j} \prod_{l=1}^{n} (1 - F_l(x + \sigma_{lj})) dx.$$

Let $\widetilde{t_{j+1}} = \widetilde{t_j} + k(t_{j+1} - t_j)$, and let us define $\widetilde{\sigma_i}(t)$ on the segment $[\widetilde{t_j}, \widetilde{t_{j+1}}]$ as follows:

$$\widetilde{\sigma_i}(t) = \begin{cases} (t - \widetilde{t_j})/k + \widetilde{\sigma_{ij}}, & \sigma_i' > 0 \text{ for } t \in [t_j, t_{j+1}] \\ \widetilde{\sigma_{ij}}, & \text{otherwise.} \end{cases} \quad (95)$$

In this case, on this segment

$$\sum_{l=1}^{n} \widetilde{\sigma_l}'(t) = 1,$$

which means that the $\widetilde{\sigma_i}$ satisfy the resource sharing constraints. By definition,

$$\widetilde{t_{j+1}} - \widetilde{t_j} = k(t_{j+1} - t_j), \quad (96)$$

and therefore for processes active on $[t_j, t_{j+1}]$ we obtain

$$\widetilde{\sigma_{i,j+1}} - \widetilde{\sigma_{ij}} = \frac{\widetilde{t_{j+1}} - \widetilde{t_j}}{k} = t_{j+1} - t_j = \sigma_{i,j+1} - \sigma_{ij}.$$

For processes idle on $[t_j, t_{j+1}]$ the same equality holds as well:

$$\widetilde{\sigma_{i,j+1}} - \widetilde{\sigma_{ij}} = 0 = \sigma_{i,j+1} - \sigma_{ij},$$

and since $\widetilde{\sigma}_i(t) = 0$ we obtain the invariant

$$\widetilde{\sigma_{ij}} = \sigma_{ij}. \tag{97}$$

The average cost for the new schedules may be represented as

$$\widetilde{E_{u_j}}(\widetilde{\sigma_1}, \ldots, \widetilde{\sigma_n}) = \int_{\widetilde{t_j}}^{\widetilde{t_{j+1}}} \left( \sum_{l=1}^{n} \widetilde{\sigma}_l' \right) \prod_{l=1}^{n} (1 - F_l(\widetilde{\sigma}_l)) dt =$$

$$\prod_{l=k+1}^{n} (1 - F_l(\widetilde{\sigma_{lj}})) \int_{\widetilde{t_j}}^{\widetilde{t_{j+1}}} \prod_{l=1}^{k} (1 - F_l((t - \widetilde{t_j})/k + \widetilde{\sigma_{lj}})) dt.$$

Substituting $x = (t - \widetilde{t_j})/k$ and using (95), (96) and (97), we obtain

$$\widetilde{E_{u_j}}(\widetilde{\sigma_1}, \ldots, \widetilde{\sigma_n}) = k \prod_{l=k+1}^{n} (1 - F_l(\widetilde{\sigma_{lj}})) \int_0^{(\widetilde{t_{j+1}} - \widetilde{t_j})/k} \prod_{l=1}^{k} (1 - F_l(x + \widetilde{\sigma_{lj}})) dx =$$

$$k \prod_{l=k+1}^{n} (1 - F_l(\sigma_{lj})) dt \int_0^{t_{j+1} - t_j} \prod_{l=1}^{k} (1 - F_l(x + \sigma_{lj})) dx =$$

$$E_{u_j}(\sigma_1, \ldots, \sigma_n).$$

From the last equation, it immediately follows that

$$E_u(\widetilde{\sigma_1}, \ldots, \widetilde{\sigma_n}) = \sum_{j=0}^{\infty} \widetilde{E_{u_j}}(\widetilde{\sigma_1}, \ldots, \widetilde{\sigma_n}) = \sum_{j=0}^{\infty} E_{u_j}(\sigma_1, \ldots, \sigma_n) = E_u(\sigma_1, \ldots, \sigma_n),$$

which completes the proof.
*Q.E.D.*

## References

Boddy, M., & Dean, T. (1994). Decision-theoretic deliberation scheduling for problem solving in time-constrained environments. *Artificial Intelligence*, *67*(2), 245–286.

Clearwater, S. H., Hogg, T., & Huberman, B. A. (1992). Cooperative problem solving. In Huberman, B. (Ed.), *Computation: The Micro and Macro View*, pp. 33–70. World Scientific, Singapore.

Dean, T., & Boddy, M. (1988). An analysis of time-dependent planning. In *Proceedings of the Seventh National Conference on Artificial Intelligence (AAAI-88)*, pp. 49–54, Saint Paul, Minnesota, USA. AAAI Press/MIT Press.

Finkelstein, L., & Markovitch, S. (2001). Optimal schedules for monitoring anytime algorithms. *Artificial Intelligence*, *126*, 63–108.

Finkelstein, L., Markovitch, S., & Rivlin, E. (2002). Optimal schedules for parallelizing anytime algorithms: The case of independent processes. In *Proceedings of the Eighteenth National Conference on Artificial Intelligence*, pp. 719–724, Edmonton, Alberta, Canada.

Gomes, C. P., & Selman, B. (1997). Algorithm portfolio design: Theory vs. practice. In *Proceedings of UAI-97*, pp. 190–197, San Francisco. Morgan Kaufmann.

Gomes, C. P., Selman, B., & Kautz, H. (1998). Boosting combinatorial search through randomization. In *Proceedings of the 15th National Conference on Artificial Intelligence (AAAI-98)*, pp. 431–437, Menlo Park. AAAI Press.

Horvitz, E. (1987). Reasoning about beliefs and actions under computational resource constraints. In *Proceedings of the Third Workshop on Uncertainty in Artificial Intelligence*, pp. 429–444, Seattle, Washington.

Huberman, B. A., Lukose, R. M., & Hogg, T. (1997). An economic approach to hard computational problems. *Science*, *275*, 51–54.

Janakiram, V. K., Agrawal, D. P., & Mehrotra, R. (1988). A randomized parallel backtracking algorithm. *IEEE Transactions on Computers*, *37*(12), 1665–1676.

Knight, K. (1993). Are many reactive agents better than a few deliberative ones. In *Proceedings of the Thirteenth International Joint Conference on Artificial Intelligence*, pp. 432–437, Chambéry, France. Morgan Kaufmann.

Korf, R. E. (1990). Real-time heuristic search. *Artificial Intelligence*, *42*, 189–211.

Kumar, V., & Rao, V. N. (1987). Parallel depth-first search on multiprocessors. part II: Analysis. *International Journal of Parallel Programming*, *16*(6), 501–519.

Luby, M., & Ertel, W. (1994). Optimal parallelization of Las Vegas algorithms. In *Proceedings of the Annual Symposium on the Theoretical Aspects of Computer Science (STACS '94)*, pp. 463–474, Berlin, Germany. Springer.

Luby, M., Sinclair, A., & Zuckerman, D. (1993). Optimal speedup of Las Vegas algorithms. *Information Processing Letters*, *47*, 173–180.

Rao, V. N., & Kumar, V. (1987). Parallel depth-first search on multiprocessors. part I: Implementation. *International Journal of Parallel Programming*, *16*(6), 479–499.

Rao, V. N., & Kumar, V. (1993). On the efficiency of parallel backtracking. *IEEE Transactions on Parallel and Distributed Systems*, *4*(4), 427–437.

Russell, S., & Wefald, E. (1991). *Do the Right Thing: Studies in Limited Rationality*. The MIT Press, Cambridge, Massachusetts.

Russell, S. J., & Zilberstein, S. (1991). Composing real-time systems. In *Proceedings of the Twelfth National Joint Conference on Artificial Intelligence (IJCAI-91)*, pp. 212–217, Sydney. Morgan Kaufmann.

Simon, H. A. (1982). *Models of Bounded Rationality*. MIT Press.

Simon, H. A. (1955). A behavioral model of rational choice. *Quarterly Journal of Economics*, *69*, 99–118.

Yokoo, M., & Kitamura, Y. (1996). Multiagent real-time-A* with selection: Introducing competition in cooperative search. In *Proceedings of the Second International Conference on Multiagent Systems (ICMAS-96)*, pp. 409–416.

Zilberstein, S. (1993). *Operational Rationality Through Compilation of Anytime Algorithms*. Ph.D. thesis, Computer Science Division, University of California, Berkeley.